\documentclass[letterpaper]{article} 
\usepackage[preprint]{aaai2027}  
\usepackage[hyphens]{url}  
\usepackage{graphicx} 
\usepackage{natbib}  
\usepackage{caption} 
\usepackage{algorithm}
\usepackage{algorithmic}

\usepackage{newfloat}
\usepackage{listings}
\DeclareCaptionStyle{ruled}{labelfont=normalfont,labelsep=colon,strut=off} 
\floatstyle{ruled}
\newfloat{listing}{tb}{lst}{}
\floatname{listing}{Listing}
\usepackage{todonotes}

\usepackage{booktabs}

\usepackage[inkscapelatex=false]{svg}
\usepackage{tabularx} 
\usepackage{amsmath}
\usetikzlibrary{arrows.meta, positioning}
\definecolor{axonkw}{HTML}{1A5276}
\definecolor{axoncmt}{HTML}{7F8C8D}
\definecolor{axonstr}{HTML}{27AE60}
\definecolor{axonopr}{HTML}{C25A0A}
\definecolor{axontype}{HTML}{6C3483}

\lstdefinelanguage{Axon}{
  morekeywords={import,export,type,if,then,else,for,do,return,yield,scope,carry,step,and,or,true,false,null},
  morekeywords=[2]{Tensor,List,Path,Dim,Int,Float,Bool,String,Any},
  sensitive=true,
  morecomment=[l]{--},
  morestring=[b]",
  morestring=[b]'
}

\title{Write Once, Run Everywhere: The Axon DSL for Shape-Safe and Framework-Agnostic LLM Architectures}
\author{
    Jacob Nielsen, 
    Danial Namazifard,
    Lukas Galke Poech,
    Peter Schneider-Kamp
}
\affiliations{
    University of Southern Denmark

Campusvej 55, 5230 Odense, Denmark\\
    jacn@imada.sdu.dk
}

\usepackage{tabularx} 
\usepackage{booktabs} 
\begin{document}

\maketitle

\begin{abstract}
The entire ecosystem of open-source language models effectively relies on a single platform. What if this platform was forced to shut down tomorrow? Implementing and maintaining efficient model definitions and translating them between different training and inference regimes is a resource-heavy task that severely limits model efficiency and portability, hindering both scaling and deployment. Here, we present Axon, a strongly typed domain-specific language with Haskell-like syntax, that enables a \emph{write-once, run everywhere paradigm} for LLM architectures. By basing collaboration on a language specification rather than a specific framework's vision, Axon fosters open cooperation and empowers researchers to implement highly specialized architectures without giving up optimization infrastructure or accepting deployment lock-in. Axon allows for concise, auditable specifications that can be automatically compiled to standalone implementations for leading frameworks: PyTorch, PyTorch with Triton, JAX, MLX and vLLM. In 467 inference benchmarking experiments on models ranging from 135M to 32B parameters, we demonstrate median speedups of $7\%$ on PyTorch, $12\%$ on PyTorch with Triton, $91\%$ on JAX, and $107\%$ on MLX, compared to the reference implementations from Transformers. When deployed as native vLLM architectures with PagedAttention and KV-cache, Axon models achieve a $58\%$ median speedup over Transformers implementations.
\end{abstract}

\section{Introduction}\label{sec:introduction}

In 1972/73, Dennis Ritchie created the C programming language \cite{ritchie1973c} which -- uniquely at the time -- combined the hardware access of low-level programming languages with the portability and syntax of high-level languages. Undoubtedly, the C language has revolutionized how we program computers.
The current situation in LLM development is reminiscent of the pre-C era. Some frameworks enable high levels of abstraction, such as the Transformers library~\cite{wolf-etal-2020-transformers}, while others are closer to hardware, such as CUDA kernels \cite{nickolls2008cuda} or FlashAttention implementations~\cite{dao2022flashattention}.
The de-facto standard implementations of LLM architectures are currently determined by rigid, monolithic platforms,  most notably the Hugging Face Hub\footnote{https://huggingface.co}, which prioritize the ease of sharing models and model definitions over efficiency.

This discrepancy between architectural transparency and execution efficiency creates a high barrier to porting models between software backends and hardware platforms: porting implementations between backends, such as PyTorch~\cite{ansel2024pytorch}, JAX~\cite{jax2018github}, and MLX~\cite{mlx2023}, is error-prone and non-trivial, creating technical debt~\cite{sculley2015hidden}.

\begin{figure}
    \centering
    \includegraphics[width=0.9\linewidth]{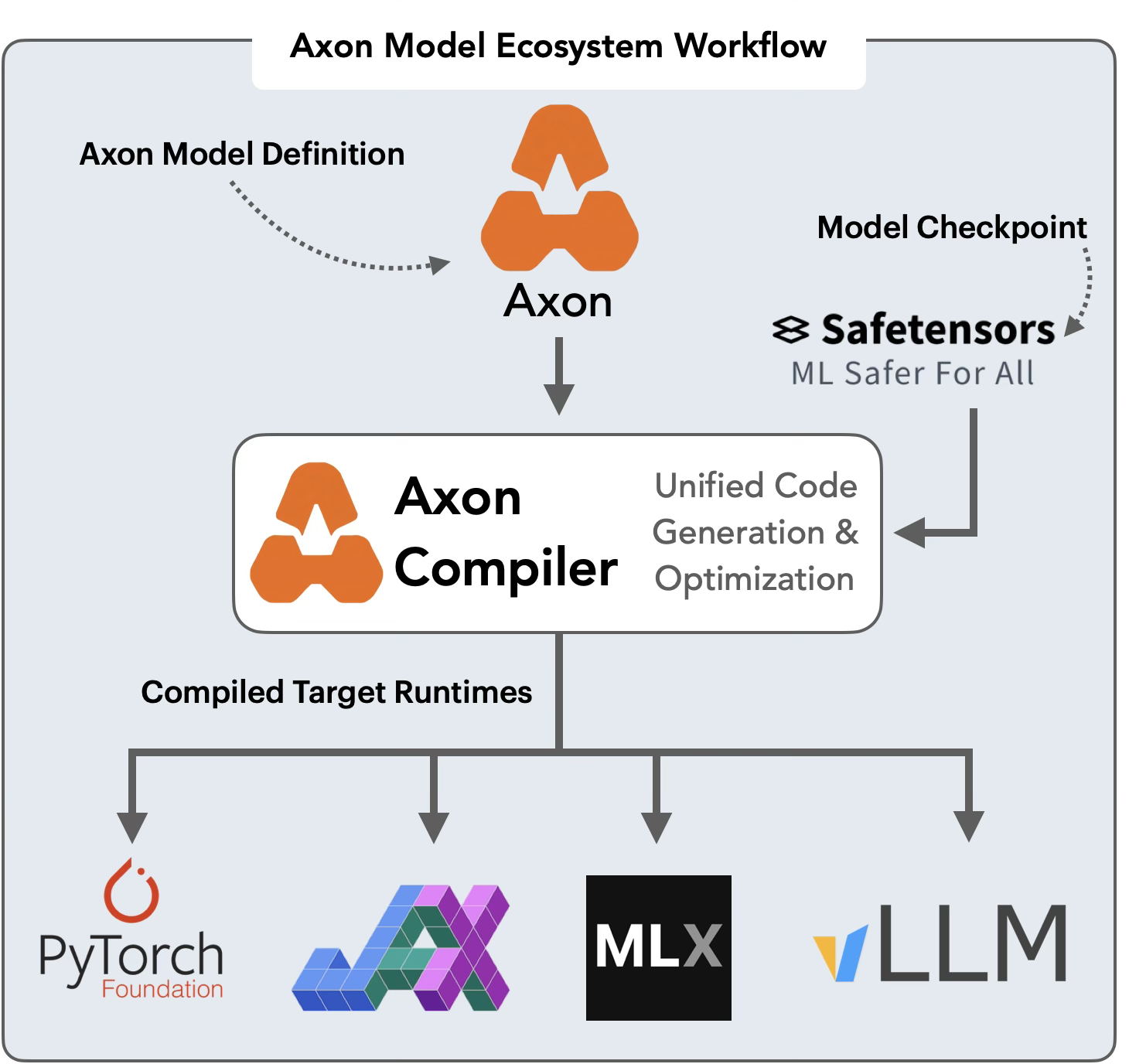}
    \caption{Write-once, run everywhere. Axon DSL compiles axon definitions (\texttt{.axon}) to standalone model definitions for PyTorch, JAX, MLX and vLLM. All that is needed is a standard \texttt{safetensors} checkpoint.}
    \label{fig:axon-compilations-backends}    
\end{figure}

Modeling code is hidden in a vast ecosystem of glue code and pipeline jungles connecting frameworks and hardware backends, freezing a system to the peculiarities of a specific package and making architectural improvements or backend transitions prohibitively expensive. Most current optimization efforts, constrained by this dependency chain, patch existing frameworks -- leading to implementation drift~\cite{sculley2015hidden}, where backend-specific optimizations are lost or omitted for compatibility with general framework conventions, and forcing researchers and practitioners to choose between (a) suboptimal performance or (b) redundantly rewriting implementations for every backend and hardware. Exacerbating this issue, the efforts of most practitioners and researchers lie in the hands of a few private companies, leading to undesirable centralization of power~\cite{ahmed2023growing, widder2024open}.

Here, we present \textbf{Axon} (Figure~\ref{fig:axon-compilations-backends}), a compact, strongly typed functional domain-specific language (DSL) that allows for minimal, standalone model definitions, and thereby reduces the dependency chains and abstraction debt~\cite{sculley2015hidden} that characterize monolithic frameworks:
Axon targets model families rather than hand-written model-specific implementations per backend, per hardware: lexical names, symbolic dimensions, parameter paths, optional configuration defaults, control-flow structure, and precise tensor types.
Furthermore, Axon removes a major source of training/serving skew identified by \citet{breck2017mltestscore} as one of the most critical yet least-frequently implemented tests for production readiness. 
Axon further mitigates the fast-integration debt \cite{ehsani2026faster} inherent in rapid LLM adoption and empowers researchers to implement and share specialized architectures without sacrificing optimization infrastructure. 

While manual porting between backends often leads to implementation drift -- where optimizations for one framework are omitted or lost in another -- Axon’s \textbf{write-once, run everywhere} paradigm facilitates integrity across the entire model lifecycle. By formalizing the model specification as an auditable and unit-testable artifact, Axon directly supports the ML Test Score criteria 
for reproducibility and ensures that architectures are inherently ready for 
diverse production environments \cite{breck2017mltestscore}.
Moreover, Axon enables a seamless generation of optimized implementations for both training and inference, currently targeting PyTorch, PyTorch with Triton, JAX, MLX, and vLLM, within a single compiler pipeline. Axon's compiler preserves the integrity of the symbolic dimensions and control-flow structures. 

In sum, Axon contributes to democratizing high-performance AI, empowering smaller entities to achieve competitive throughput while enabling large-scale operators to optimize hardware utilization and reduce their environmental footprint.
Our contributions are as follows: 
\begin{itemize}
    \item A strongly-typed domain-specific language, Axon, for authoring backend-agnostic neural (language) models for cross-backend materializations.
    \item A compiler pipeline for seamless compilations of optimized standalone implementations for over 204 models spanning 60 families. 
    \item Benchmarks of these checkpoints across five backends displaying improved or at least competitive results in terms of throughput, while maintaining top-1 token parity across backends. 
    \item A deployment path to multiple backends based on a checkpoint and the Axon model definition.
\end{itemize}

\section{Related Work}\label{sec:related-work}
The ecosystem for optimizing deep neural networks spans high-level API abstractions to low-level hardware primitives. We broadly categorize existing work into five layers: (i) deep learning backends, (ii) intermediate representations, (iii) compilers and optimizers, (iv) kernel generation, and (v) specialized inference frameworks.

\paragraph{Deep Learning Backends.} At the highest level, we have well known backends such as PyTorch~\cite{ansel2024pytorch}, TensorFlow~\cite{Abadi_TensorFlow_Large-scale_machine_2015}, and JAX~\cite{jax2018github} providing the tensor operations from which model definitions and their execution are built with the \texttt{backward} pass typically derived automatically. While PyTorch and TensorFlow mainly rely on eager execution, the industry has shifted towards just-in-time (JIT) compilations for performance. More recently, specialized backends, such as Tinygrad~\cite{tinygrad} and MLX~\cite{mlx2023}, emphasize minimalism and hardware-specific optimizations to reduce overhead often introduced by the general-purpose frameworks.

\paragraph{Intermediate Representations (IR).} To map high level abstractions to hardware-specific code, IR frameworks maintain a structured representation of computational graphs. ONNX~\cite{onnx} serves as a cross-platform standard for model exchanges, while MLIR (Multi-level IR)~\cite{9370308} and LLVM~\cite{1281665} provide modular infrastructure for dialect-specific optimizations. StableHLO~\cite{stablehlo} bridges JAX and TensorFlow, lowering high-level operations into a representation that compilers can efficiently optimize. IREE~\cite{iree} extends this with an end-to-end pipeline from MLIR to executable code across diverse devices. 

\paragraph{Compilers and Optimizers.} Once represented in an IR, compilers optimize the graph through techniques such as operator fusion, memory planning, and constant folding. Accelerated Linear Algebra (XLA)~\cite{50530} does this for JAX and TensorFlow, while the \texttt{torchinductor} is the primary backend for PyTorch 2.0 \cite{ansel2024pytorch}. Hardware specific optimizers like NVIDIA's TensorRT~\cite{nvidia2026tensorrt} and Intel's OpenVINO~\cite{intelopenvino} maximize throughput for specific chipsets. 
Apache TVM~\cite{chen2018tvm} provides general-purpose automated optimization, leveraging machine learning to find efficient operator implementations. 

\paragraph{Kernel Generation.} The lowest level of optimization happens at the kernel level, where mathematical operations are mapped to hardware threads/warps in a tensorized formulation. Native libraries, such as NVIDIA's CUTLASS~\cite{cutlass} and cuDNN~\cite{chetlur2014cudnn}, and the cross-platform oneDNN~\cite{onednn}, provide highly tuned primitives.
A fast-growing trend towards higher-level, programmable kernel generation has followed: Triton~\cite{tillet2019triton} compiles high-level Python to efficient GPU kernels, DITRON targets distribution-aware kernels~\cite{zheng2026ditron}, and Mojo\footnote{https://mojolang.org/docs/} bridges Python's usability with the performance of C++ and CUDA. LLMs now also power this space, e.g.\ as part of the AutoKernel project\footnote{https://github.com/rightnow-ai/autokernel}.

\paragraph{Specialized Inference Frameworks.} Specialized inference frameworks wrap prior layers of optimization for deployments and serving. vLLM increases serving throughput through PagedAttention, a virtual-memory-inspired KV-cache manager~\cite{kwon2023efficient}. For Apple Silicon, emerging projects like MLX-vLLM~\cite{barrios2025vllmmlx} and Rapid MLX~\cite{chai2026rapidmlx} adapt these high-performance serving techniques to the MLX ecosystem. 

\paragraph{Summary.} A common thread across these five layers is that each is specialized to a particular software or hardware stack. Backends expose framework-specific model (tensor) APIs. IRs define machine-exchange formats rather than authoring languages, compilers optimize within a single stack, kernel generation targets specific hardware, and inference frameworks serve one runtime. Portability across stacks at any level is achieved by manually re-expressing implementations. Axon occupies a complementary layer \emph{above} these: a human-authored, strongly typed specification language with symbolic dimensions, from which a single Graph IR is lowered to standalone native implementations for PyTorch, Triton-accelerated PyTorch, JAX, MLX, and vLLM. By enforcing one Graph IR contract consumed by every backend, Axon ensures that a definition-level optimization cannot be present in one framework's port yet silently omitted in another. 

\section{The Axon Language}\label{sec:axon-language}
Axon is a compact functional language for describing neural networks independently of any particular implementation or runtime backend, producing standalone model definitions.
A more detailed description of the language including Backus-Naur form is provided in the technical supplement.

\paragraph{Values.}
The first-class domain objects in Axon include tensors, symbolic dimensions, and paths into checkpoint state-dicts and configurations. The language is intentionally designed to describe the mathematical structure of the neural network rather than the mechanics of a specific implementation in a particular language or backend.

\paragraph{Types.}
The tensor types carry shapes comprised of symbolic dimensions. A type such as \lstinline{x :: Tensor[B,S,D]} states that a value \lstinline{x} is a tensor \lstinline{BxSxD}, representing a batch of sequences of hidden-state vectors. 
Symbolic dimensions are part of the static semantics, participating in type checking, primitive typing rules, and the Graph IR metadata.
As a consequence, Axon definitions can express reusable shape relationships without hard-coding checkpoint-specific sizes. Symbolic dimensions can operate as values in expressions, reducing the retrieval of the sequence length of a tensor \lstinline{x} from an expression such as \lstinline{x.shape[-2]} to simply \lstinline{S}.

\paragraph{Syntax.}
Axon's syntax is Haskell-inspired. Definitions use signatures, arrow types, optional parameters, pipes, and a do-style statement form. As an example, consider a pre-norm attention sub-layer, in which the same normalized hidden state \lstinline{h} is passed as query, key, and value, and \lstinline{@ln} indicates that the RMS-norm parameters live under the current scope's \lstinline{ln} key in the checkpoint's state-dict.

\begin{lstlisting}[basicstyle=\ttfamily\scriptsize]
block :: Tensor[B,S,D] -> ?Tensor[B,K] -> Tensor[B,S,D]
block x attn_mask = do
  h <- NN.rmsnorm@ln x
  a <- Attention.attention h h h attn_mask
  return x + a
\end{lstlisting}
Scoped parameter paths are one of several surface conveniences that will be normalized before flattening during compilation. 

\paragraph{Functional semantics.} Semantically, Axon definitions are functions over explicit or implicit arguments, configuration lookups, and parameter-path reads. A bind statement names the values produced by an expression. Except for ternary branch expressions, all expressions are eager, i.e., an argument is evaluated before the callee receives it as a value.

\paragraph{Model paths.} Model parameters and config values are accessed through values of type \lstinline{Path}. Relative path values are denoted using ``\lstinline{@}'': \lstinline{NN.rmsnorm@ln} from our example refers to a path scoped by the current lexical context. Absolute paths are denoted using ``\lstinline{@@}'': \lstinline|NN.rmsnorm@@'layers.{i}.attn.ln'| refers to checkpoint locations directly, with \lstinline{i} being instantiated from the runtime context. Whenever possible, relative paths are normalized into absolute paths by the compiler.

\paragraph{Reusable libraries.} Primitive Axon operations, such as the tensor operation \lstinline{_where} or parameter retrieval \lstinline{_param}, are implemented as internal definitions prefixed by an underscore. They are exposed through lightweight wrappers as part of reusable library modules, such as \lstinline{NN}, \lstinline{Tensor}, \lstinline{Attention}, \lstinline{Masking}, \lstinline{MoE}, \lstinline{Cache}, and \lstinline{Positions}. In addition to exposing primitive operations, these modules define reusable higher-level generic model-independent operations such as \lstinline{Attention.attention} and \lstinline{Cache.prepare}.

\paragraph{Semantic core.}
The full Axon DSL language offers expressiveness through library imports, default arguments, nested expressions, path scoping, operation pipes, and other syntactic sugar. Full Axon DSL can be desugared to a small semantic core where library references, default arguments, and paths are resolved, and nested expressions are flattened and fully typed, at the subexpression level. This semantic core remains valid Axon DSL but has a one-to-one correspondence to Axon's Graph IR.

\section{The Axon Compiler}\label{sec:axon-compiler}
The Axon compiler comprises three main phases: (1) desugaring of full Axon DSL into its semantic core; (2)  optimization at the Graph IR level; and (3) lowering of the Graph IR to backend-specific code.
In the technical supplement, we walk an exemplary small definition through all phases.

\paragraph{Phase 1: Desugaring.}
The Axon DSL is defined by an EBNF grammar that is operationalized using Lark \cite{lark}. The parser transforms valid Axon DSL into an abstract syntax tree (Axon AST), which serves as the representation for Phase 1 and can be dumped to Axon DSL and reparsed. Every desugaring stage of Phase 1 receives an Axon AST and outputs an Axon AST, typically guaranteeing additional properties: (a) the resolver stage inlines library definitions imported and referenced, guaranteeing a closure property stating that all non-primitive operations are expressed as definitions in the AST; (b) the flattening stage flattens nested expressions, guaranteeing a flatness property stating that definitions consist of a flat sequence of operations; (c) the type checker stage infers types from type signatures, type ascriptions, and the type rules of primitive operations, guaranteeing a consistently typed Axon AST. 
The desugaring phase also contains dead code elimination, constant folding, and other straightforward code normalizations.

\paragraph{Phase 2: Graph optimizations.}
The closed flat typed Axon AST can be viewed as an instance of a static single assignment (SSA) graph~\cite{cytron1991ssa}. In this phase, the Axon compiler converts this semantic core to the Graph IR, which is then iteratively optimized using a number of stages: \emph{(a) definition inlining} replaces wrapper definitions and other low-complexity definitions by integrating their sequence of operations into their call sites, enabling, for example, the inlining of primitive wrapper definitions; \emph{(b) generic graph rewriting} replaces known patterns of operations by equivalent patterns of operations based on matching of primitive operational provenance, enabling, for example, the replacement of loops iterating over experts in an MoE by a more efficient stacked-experts formulation; and \emph{(c) backend-specific graph rewriting} replaces known patterns of operations by backend-specific primitives, enabling, for example, the use of optimized attention implementations and kernels.

\paragraph{Phase 3: Lowering.}
The optimized Graph IR contains definitions as assignment sequences, calls to definitions, and generic and backend-specific primitive operations. The lowering phase is backend-specific and transpiles the Graph IR into one of the currently targeted backends: PyTorch, Triton-accelerated PyTorch, MLX, JAX, and vLLM. Each of the separate code generators needs to be able to implement the control flow and the primitive operators represented by the Graph IR. In addition, the code generators also need to implement \texttt{forward} and \texttt{generate} methods. To achieve competitive performance, compilation-oriented backends typically supply scaffolding that ensures static cache allocations.

\section{Results}\label{sec:results}
We now present a series of results from our Axon-derived models across PyTorch, JAX, MLX and vLLM backends. The results will establish that Axon-derived models yield performance parity, and often outperform implementations in existing frameworks. We further show that Axon-derived models yield near-identical training behavior. All figures can be found in larger size in the technical supplement.

\begin{figure}[t]
  \centering
  \includesvg[width=\columnwidth]{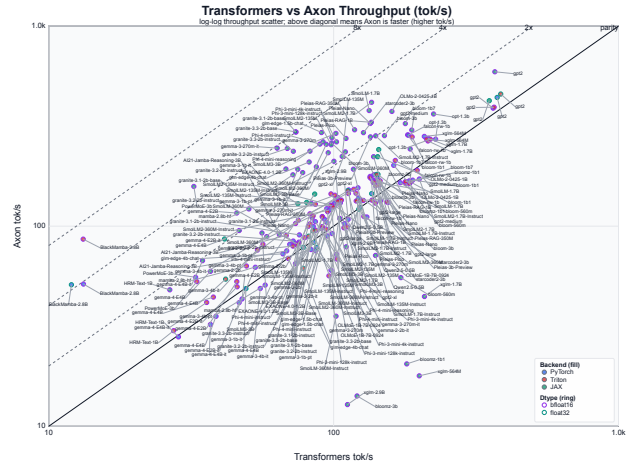}
  \caption{Autoregressive generation performance with decoder-only models with up to $4B$ parameters. Log-log plot where the diagonal line indicates equal performance. Points above the parity line indicate Axon is faster, color-fill denotes backends and border-line denotes dtype.}
  \label{fig:bench4b:autoreg}
\end{figure}

\paragraph{Experimental Setup}
MLX backend experiments on $<1B$ models were executed on Apple Silicon M3 Max, 36GB RAM with MLX \texttt{0.32.0}. We enabled compilation and allowed a max generation length of 256 tokens. PyTorch, Triton-enhanced PyTorch, JAX and vLLM experiments were conducted on an NVIDIA B200 GPU with 180 GB HBM3, using PyTorch \texttt{2.11.0}, Triton \texttt{3.6.0}, JAX \texttt{0.10.2}, and Transformers \texttt{5.10.0.dev0}. Models that can natively be compiled without errors and all Axon models are compiled. All models ran with $1$ generation-warmup step, $3$ repetitions, and a max generation length of 128 tokens. During these experiments we exercise the \texttt{generate()} method in the models. Both the Axon-compiled model and the Transformers baseline use \texttt{torch.compile} and equivalent JIT compilations for Axon using the \texttt{--compile-axon}, \texttt{--compile-hf}, \texttt{--optimize-graph} flags. The primary precision is \texttt{bfloat16}, with \texttt{float32} fallback for pairs that fail top-1 token parity with the Transformers baseline.

\paragraph{Measures}
On generation benchmarks, throughput is calculated from per-run log output as $\text{generated tokens} / \text{wall time}$, where $\text{generated tokens}$ is the actual number of tokens produced before the end-of-sequence token, rather than the maximum generation length cap. All comparisons use a \emph{runtime ratio}
$\rho = t_{\text{Axon}} / t_{\text{Transformers}}$ where $t_{\text{Axon}}$ and $t_{\text{Transformers}}$ are the wall-clock times for the Axon-derived model and the reference (Transformers) implementation under identical inputs, precision, and compilation settings. A value $\rho < 1$ indicates that Axon is faster, while $\rho = 1$ represents parity.

\begin{table}[t]
\centering
\caption{Per-backend comparison of \textbf{Axon's autoregressive generation performance with decoder-only models} against PyTorch, Triton, and JAX.
  ``Axon $\le 1\times$'' = count (and \%) of checkpoints where Axon is faster than Transformers. ``Axon $> 1\times$'' = count
  (\%) of checkpoints where Axon is slower. Median and mean report runtime ratio (lower means Axon is faster).
}\label{tab:4b-backend-comparison}\label{tab:32b-backend-comparison}
\resizebox{\columnwidth}{!}{
\begin{tabular}{lccccc}
\toprule
Backend & Checkpoints & Axon $\le 1\times$ & Axon $> 1\times$ & Median ratio & Mean ratio \\
\midrule
\multicolumn{6}{c}{\textbf{<4B}}\\
PyTorch & 76 & 48 (63\%) & 28 (37\%) & 0.903 & 0.925 \\
Triton  & 75 & 60 (80\%) & 15 (20\%) & 0.843 & 0.878 \\
JAX     & 74 & 64 (86\%) & 10 (14\%) & 0.481 & 1.122 \\
\midrule
\textbf{<4B total} & \textbf{225} & \textbf{172 (76\%)} & \textbf{53 (24\%)} & \textbf{0.804} & \textbf{0.974} \\
\midrule
\midrule
\multicolumn{6}{c}{\textbf{4--32B}}\\
PyTorch & 87 & 55 (63\%) & 32 (37\%) & 0.987 & 0.980 \\
Triton  & 87 & 67 (77\%) & 20 (23\%) & 0.924 & 0.931 \\
JAX     & 68 & 55 (81\%) & 13 (19\%) & 0.589 & 0.981 \\
\midrule
\textbf{4--32B total} & \textbf{242} & \textbf{177 (73\%)} & \textbf{65 (27\%)} & \textbf{0.908} & \textbf{0.963} \\
\bottomrule
\end{tabular}
}
\end{table}

\begin{table}[t]
\centering
\caption{Per-backend comparison of \textbf{Axon's Forward performance with Encoder-Only and Encoder-Decoder Models} against PyTorch, Triton, and JAX backends.
  ``Axon $\le 1\times$'' = count (and \%) of checkpoints where Axon is faster than Transformers. ``Axon $> 1\times$'' = count
  (\%) of checkpoints where Axon is slower. Median and mean report runtime ratio (lower means Axon is faster).
} 
\label{tab:backend-comparison-forward}\label{tab:backend-comparison-forward-32b}
\resizebox{\columnwidth}{!}{
\begin{tabular}{lccccc}
\toprule
Backend & Checkpoints & Axon $\le 1\times$ & Axon $> 1\times$ & Median ratio & Mean ratio \\
\midrule
\multicolumn{6}{c}{\textbf{$\leq$4B}}\\
PyTorch & 26 & 15 (58\%) & 11 (42\%) & 0.883 & 1.786 \\
Triton  & 26 & 9 (35\%)  & 17 (65\%) & 1.332 & 2.169 \\
JAX     & 26 & 7 (27\%)  & 19 (73\%) & 1.079 & 1.260 \\
\midrule
\textbf{$\leq$4B total} & \textbf{78} & \textbf{31 (40\%)} & \textbf{47 (60\%)} & \textbf{1.084} & \textbf{1.738} \\
\midrule
\midrule
\multicolumn{6}{c}{\textbf{4--32B}}\\
PyTorch & 20 & 13 (65\%) & 7 (35\%)  & 0.990 & 1.506 \\
Triton  & 20 & 8 (40\%)  & 12 (60\%) & 1.133 & 1.546 \\
JAX     & 14 & 0 (0\%)   & 14 (100\%) & 3.747 & 3.356 \\
\midrule
\textbf{4--32B total} & \textbf{54} & \textbf{21 (39\%)} & \textbf{33 (61\%)} & \textbf{1.175} & \textbf{2.001} \\
\bottomrule

\end{tabular}
}
\end{table}

\subsection{Sub-$4B$ Models}
We benchmark 91 checkpoints spanning 71 Axon model definitions up to 4B parameters, across 26+ architecture families including GPT, Llama, Mistral, Qwen, Gemma, SmolLM, T5, BART, BERT, and Mamba -- measured across three backends (PyTorch, Triton, JAX). Of these, 68 are decoder-only language models tested via autoregressive generation (Figure~\ref{fig:bench4b:autoreg}), and 23 are encoder-only or encoder-decoder (seq2seq) models tested via forward passes (Table~\ref{tab:backend-comparison-forward} and plots in the technical supplement).

Each autoregressive generation produces at least 8 tokens across the 225 checkpoints (200 BF16, 25 FP32), with a median of 123 tokens, where $91\%$ generated at least 100 tokens. Nine checkpoints failed BF16 top-1 token parity due to precision noise and were re-run in FP32, achieving $100\%$ top-1 token parity across all 303 checkpoints. Forward benchmarks are reported in wall-time (ms), as mean of 3 repeated measurements excl.\ a warm-up pass, covering 78 checkpoints across 23 encoder/seq2seq families (69 BF16, 9 FP32).

Autoregressive Axon-derived models are comparable or faster than Transformers for 76\% of the 225 checkpoints, with a median speed ratio of $0.804\times$ and a mean of $0.974\times$---a typical Axon model is $24\%$ faster. We report a per-backend breakdown in Table \ref{tab:4b-backend-comparison}. Most checkpoints run on all three backends; a small number did not complete on every backend, yielding 76 PyTorch, 75 Triton, and 74 JAX generate checkpoints (bloomz-560m generated fewer than 5 tokens and was excluded; BlackMamba-2.8B did not complete on JAX).
JAX achieves the best overall generate throughput with a median of $0.481\times$ and $86\%$ at or below parity, though it also has the highest variance with a mean of $1.122\times$, driven by outliers on small \texttt{BLOOM} and \texttt{XGLM} models. Triton is the most consistent with $80\%$ at or below parity and a mean of $0.878\times$. PyTorch is competitive but has more outliers above $1\times$.
For encoder-only and encoder-decoder models (Table \ref{tab:backend-comparison-forward}), performance is generally weaker, particularly for the \texttt{T5} and \texttt{mT5} encoder-decoder models. The worst forward outliers are t5-small on Triton ($9.2\times$), t5-base on PyTorch ($4.1\times$), and mt5-large on Triton ($3.7\times$). PyTorch is the best forward backend with $58\%$ at or below parity and a median of $0.883\times$, while Triton and JAX struggle with the short forward pass.

\subsection{Models between $4B$ and $32B$}
\begin{figure}
  \centering
  \includesvg[width=\columnwidth]{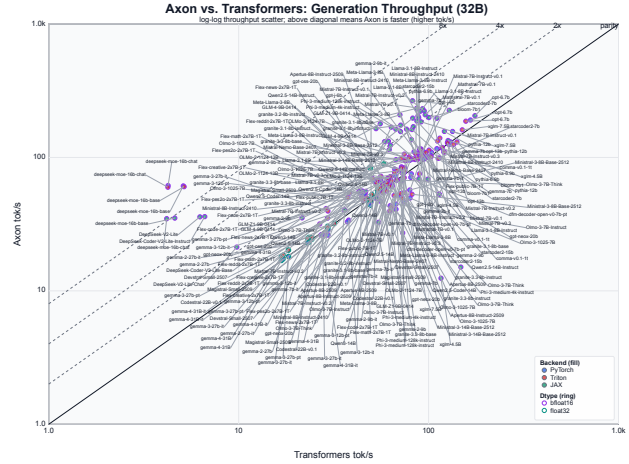}
  \caption{Benchmarking Autoregressive models between $4B$ and $32B$ models. 
  Log-log plot where the diagonal parity line indicates equal performance. Generally, Axon produces comparable or superior throughput on the bigger models.
  }
  \label{fig:bench32b:autoreg}
\end{figure}

We benchmark 93 checkpoints spanning 47 Axon model definitions between 4B and 32B parameters, yielding 242 generate checkpoints (205 BF16, 37 FP32; Figure~\ref{fig:bench32b:autoreg}, Table~\ref{tab:32b-backend-comparison}) and 54 forward-pass checkpoints (Table~\ref{tab:backend-comparison-forward-32b}, figure in technical supplement). Most checkpoints run on all three backends; JAX coverage is narrower at 68 generate checkpoints, as some models exceeding 20B parameters did not complete due to memory constraints -- we only benchmarked models fitting within a single GPU.

Each generate run produced at least 8 tokens, with a median of 122 tokens, where $94\%$ generated at least 100 tokens. 16 pairs failed BF16 top-1 parity across nine unique checkpoints, of which eleven were fixed by FP32. \texttt{Flex-code-2x7B-1T} (JAX), \texttt{Flex-news-2x7B-1T} (all backends), and \texttt{Gemma-7b} (JAX) did not meet the top-1 token parity criterion for all tested prompts. Forward benchmarks cover 54 checkpoints across 9 seq2seq families, reported in wall-time (ms) as the mean of 3 repeated measurements excluding 1 warm-up pass.

In Figure \ref{fig:bench32b:autoreg}, Axon is at or below Transformers' latency for $73\%$ of checkpoints with a median speed of $0.908\times$ and a mean of $0.963\times$. We report a per-backend breakdown for generative models in Table \ref{tab:32b-backend-comparison}. JAX achieves the best generation throughput with a median of $0.589\times$ and $81\%$ at or below parity, though with higher variance than in the $\leq$4B tier. Triton is the most consistent at $77\%$ at or below parity. PyTorch is competitive with $63\%$ at or below parity. Notably, \texttt{DeepSeek-MoE (16B)} runs $7$--$14\times$ faster across Axon backends, vastly outperforming Transformers' implementation. The worst generate outliers are concentrated on JAX for XGLM models: \texttt{xglm-7.5B} and \texttt{xglm-4.5B}.

Forward performance is weak across all backends, particularly for large encoder-decoder models like \texttt{T5-11B}, \texttt{MT5-XXL}, \texttt{t5gemma} where the encoder forward path has high kernel launch overhead. JAX is universally slower on forward (100\% above parity, median $3.747\times$). PyTorch is the best forward backend with $65\%$ at or below parity and a median of $0.990\times$.

\begin{figure}[t]
  \centering
  \includesvg[width=\columnwidth]{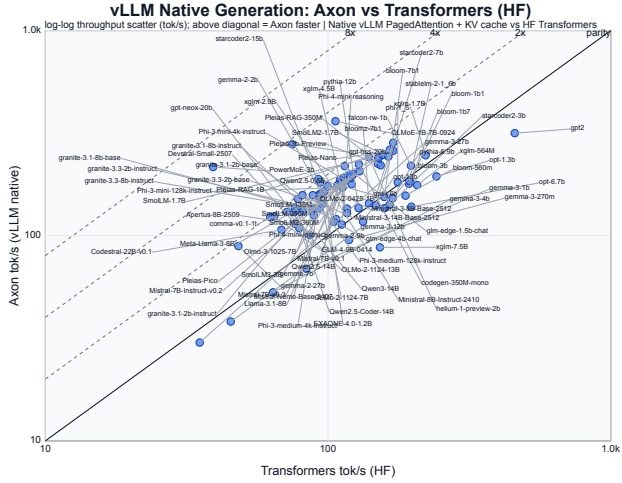}
    \caption{vLLM: Axon (vLLM native) vs. Transformers generation throughput on 88 checkpoints. Points above the parity line indicate Axon is faster. $74\%$ fall above.}
  \label{fig:vllm}
\end{figure}

\begin{table}[t]
\centering
\caption{
  Autoregressive generation comparison of Axon (vLLM native) against Transformers.
  ``Axon $\le 1\times$'' = number (and \%) of checkpoints where Axon is faster than Transformers. ``Axon $> 1\times$'' = number
  (and \%) of checkpoints where Axon is slower than Transformers. Median and mean report Axon-to-Transformers runtime ratio (lower is faster).
}
\label{tab:vllm-native-comparison}
\resizebox{\columnwidth}{!}{
\begin{tabular}{lccccc}
\toprule
Model size & Checkpoints & Axon $\le 1\times$ & Axon $> 1\times$ & Median ratio & Mean ratio \\
\midrule
Small ($\le$4B) & 54 & 38 (70\%) & 16 (30\%) & 0.656 & 1.430 \\
Large (4B--32B) & 34 & 27 (79\%) & 7 (21\%) & 0.609 & 2.889 \\
\midrule
\textbf{Total} & \textbf{88} & \textbf{65 (74\%)} & \textbf{23 (26\%)} & \textbf{0.631} & \textbf{1.994} \\
\bottomrule
\end{tabular}
}
\end{table}
\subsection{vLLM}
In Figure \ref{fig:vllm} and Table \ref{tab:vllm-native-comparison}, we report on Axon derived native vLLM models. We test 88 checkpoint-backend pairs covering models from 135\,M to 30\,B parameters.  Axon-compiled vLLM is faster than HF Transformers in 65 of 88 cases (74\%), with a median runtime ratio of 0.631. At the median, Axon completes generation in 63\% of the time taken by HF Transformers ($1.6\times$ speedup).

\begin{figure}
  \centering
  \includesvg[width=\columnwidth]{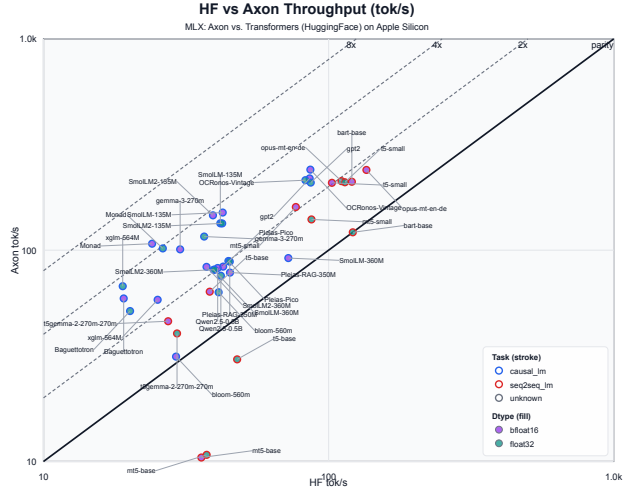}
  \caption{MLX. Benchmarking conventional HF models against Axon derived standalone model definitions. Axon yields some considerable speed-ups across the board.}
  \label{fig:mlx-bench}
\end{figure}

\begin{table}[t]
\centering
\caption{Breakdown of Axon performance by precision, model type, and sequence length.
``Axon $\le 1\times$'' = number (and \%) of checkpoints where Axon is faster than Transformers. ``Axon $> 1\times$'' = number
  (and \%) of checkpoints where Axon is slower than Transformers. Median and mean report Axon-to-Transformers runtime ratio (lower = Axon is faster).
}
\label{tab:mlx-bench:group-comparison}
\resizebox{\columnwidth}{!}{
\begin{tabular}{lccccc}
\toprule
Group & Points & Axon $\le 1\times$ & Axon $> 1\times$ & Median ratio & Mean ratio \\
\midrule
BF16 & 63 & 61 (97\%) & 2 (3\%) & 0.423 & 0.500 \\
FP32 & 63 & 58 (92\%) & 5 (8\%) & 0.503 & 0.588 \\
\midrule
causal\_lm  & 84 & 82 (98\%) & 2 (2\%)  & 0.390 & 0.437 \\
seq2seq\_lm & 42 & 37 (88\%) & 5 (12\%) & 0.574 & 0.752 \\
\midrule
len=64  & 42 & 42 (100\%) & 0 (0\%)  & 0.417 & 0.448 \\
len=128 & 42 & 39 (93\%)  & 3 (7\%)  & 0.483 & 0.525 \\
len=256 & 42 & 38 (90\%)  & 4 (10\%) & 0.494 & 0.652 \\
\midrule
\textbf{Total} & \textbf{126} & \textbf{119 (95\%)} & \textbf{7 (5\%)} & \textbf{0.483} & \textbf{0.544} \\
\bottomrule
\end{tabular}
}
\end{table}

\subsection{MLX}
In Figure \ref{fig:mlx-bench}, we report the ratios between Axon and Transformers for the MLX backend (MPS). All 21 models have $100\%$ parity in FP32, but 5 models have mismatches in BF16: \texttt{bloom-560m}, \texttt{smollm-135m}, \texttt{smollm2-135m}, \texttt{mt5-small}, \texttt{mt5-base}. Axon is at or below Transformers' latency for $94.4\%$ of the 126 checkpoints, with a median runtime ratio of $0.483\times$ and a mean of $0.544\times$ (Table~\ref{tab:mlx-bench:group-comparison}).
The decoder-only models achieve the best throughput with $98\%$ at or below parity. The fast forward path using MLX's \texttt{fast} module with Metal kernels gives \texttt{Monad} and \texttt{Gemma3-270M} the largest speedups, at roughly $4.5\times$ faster. We see that performance narrows at longer contexts. 

\subsection{Training with Axon Derived Models.}
We conducted experiments on the PyTorch backend, comparing with the Transformer's implementation as a reference. We experiment with Gemma-3 270M on the ArXiv Summarization dataset\footnote{https://huggingface.co/datasets/ccdv/arxiv-summarization}, because it is a well known task. The experiments are carried out on an NVIDIA RTX A6000 GPU, with a batch size of $4$ and a learning rate of $1e-4$. We report the loss curve for 2000 steps in the technical supplement. We compiled both models with \texttt{torch.compile} with 61s and 42s first step time for Axon and Transformers models, respectively, but with average step-time on $120.3$ms for Axon against $133.0$ms for Transformers, making Axon $9.6\%$ faster. Both models had a final loss of $2.506$, with a numerical instability of $3.74\times10^{-4}$. The only modification on the Axon-derived PyTorch model was enabling \texttt{requires\_grad}.

\section{Discussion}
The C language resolved a tension between hardware access and portability by fixing a single specification that compilers could lower to many machines.
Axon targets an analogous situation in LLM development, where architectural transparency and dependency chains are locked inside monolithic frameworks.
Across five runtime backends, we have demonstrated the potential of the Axon language under the ``write once, run everywhere''-principle: a single specification achieves competitive execution times on over 204 model checkpoints across 60 model families. 

Moreover, across 225 autoregressive checkpoints up to 4B and 242 checkpoints between 4B and 32B, Axon-derived models are as fast or faster than the Transformers baseline for $76\%$ and $73\%$ of the checkpoints, with median runtime ratios of $0.804\times$ and $0.908\times$, respectively. On the MLX backend on Apple Silicon, Axon reaches $94.4\%$ at or below parity in inference latency, median of $0.483\times$ (twice as fast) over 126 checkpoints. These numbers indicate that a single \texttt{.axon} definition is competitive across five software runtimes and two hardware vendors.

We achieve top-1 token parity on almost all models, though some required an FP32 fallback. Axon-derived models generally provide comparable or better performance across backends, though some remain slightly slower than the baseline. We believe future maturing of the compiler, especially the lowering mechanisms, will close this gap. We suspect kernel-launch overhead dominates the relatively short generation time for these slower models. See technical supplement. 

\paragraph{Portability and Shape Safety.}
A single Axon definition yields successful results across four backends from one Graph IR contract.
Because all four backends are generated from one shared IR rather than maintained as independent handwritten ports, a definition-level optimization cannot be omitted in one framework's implementation while present in another. Remaining differences are pushed down to backend-specific lowering and kernel execution.
The shape safety promised by Axon's strong type system is not directly measured by the throughput benchmarks, but the results are evidence of it: top-1 token parity across backends with distinct tensor libraries and kernel implementations is only possible because symbolic dimensions and the tensor-type contract are preserved end-to-end.
From surface Axon through the Graph IR to each backend, a shape mismatch rejected by the typechecker never reaches execution.

\paragraph{Axon for Training.} We demonstrate a training scheme for Axon-derived models: a model derived to PyTorch exhibits the same training behavior as the corresponding Transformers implementation. Already on a 270M model the Axon model yielded roughly $10\%$ faster step-times. 
We expect the impact to be larger on bigger models, and leave this for future work.
Axon therefore supports both training and inference from the same Graph IR contract, making it possible to train on one backend and serve on another.

\paragraph{Limitations.}
This work has several limitations:
Although Axon operates on tensor level and facilitates arbitrary neural network architectures, our experiments are currently limited to language models.
The main autoregressive and forward benchmarks are done on a single GPU type. MLX results are likewise on a single Apple Silicon configuration. Hardware variance is therefore not characterized. Some checkpoints required the FP32 fallback to meet top-1 token parity, indicating room to improve code generation for specific operations.
Forward performance is weak across backends particularly for large encoder-decoder models, calling for future work. We present no ablation isolating the AST and Graph IR optimization passes, making it hard to attribute speed-ups to specific compiler stages.
Lastly, we have experimented with using LLM-based coding assistants to convert Transformers model definitions to Axon; we carefully verified functional validity, though the code structure and maintainability of a few of the LLM-generated Axon definitions could be improved. Nevertheless, we regard it as promising that LLMs are able to generate functionally valid code in our newly proposed Axon language.

\paragraph{Practical Impact.}
LLMs are usually formulated as a mixture of Python modules and classes, checkpoint-specific configuration conventions and tensor state dictionaries.
Axon changes that, enabling a platform for open collaboration with easily hackable, inspectable implementations free of dependency chains controlled by a framework's vision.
This enables both cross-platform and platform-specific optimization in the search for maximum performance per Watt and per byte of VRAM.
We hope this better democratizes and secures the community's contributions than the current state, learning and applying important lessons from the history of computer science and the story of the C language.
 
\section{Conclusion}
We have introduced the Axon DSL, demonstrating that shape-safe, framework-agnostic compilation from Axon definitions to standalone implementations is a viable path toward open cooperation and greater freedom for researchers and practitioners.
Axon achieves better or competitive throughput on most models across PyTorch, Triton-enhanced PyTorch, JAX, MLX, and vLLM backends. 

\subsubsection{Future Work}\label{sec:discussion:futurework}
This work opens several directions: (i)~natively distributed tensors for finer control over communication, (ii)~third-party kernel injection from projects such as Unsloth\footnote{https://github.com/unslothai/unsloth} and Liger-kernel\footnote{https://github.com/linkedin/Liger-Kernel} as opt-in graph optimizations, and (iii)~additional backends such as Vulkan and CoreML, extending write-once, run-everywhere reach to edge devices.

\bibliography{aaai2027}

\section{Technical Supplement}\label{sec:appendix}

\subsection{Terminology}
\label{sec:appendix:terminology}
Throughout the paper we distinguish three levels of granularity. An \textbf{Axon module} is a single \texttt{.axon} source file with a number of Axon definitions, typically describing a model family or a model for a particular checkpoint, for example \texttt{generic-llama3.axon}. A \textbf{checkpoint} is one set of pretrained weights that a definition can be materialized against, for example \texttt{meta-llama/Llama-3.2-1B}; a single definition typically covers several checkpoints, which it declares through a \texttt{CHECKPOINTS} pragma. A \textbf{benchmark run} is one measurement of one checkpoint on one backend at one precision, and, for the MLX experiments, at one generation length.
Because each checkpoint is measured on several backends, the per-backend counts reported in the tables exceed the number of distinct checkpoints.

\subsection{Extended Results}
This section reproduces the encoder-only and encoder--decoder benchmarks, which are summarized
in the main text, and provides full-width versions of the remaining benchmark in Figures \ref{fig:app:bench4b:autoreg}, \ref{fig:app:bench32b:autoreg}, \ref{fig:app:vllm}, \ref{fig:app:mlx}, \ref{fig:bench4b:enc-and-enc-dec}, and \ref{fig:bench32b:enc-and-enc-dec}.

\subsection{The Axon Language}\label{sec:appendix:axon-language}

\subsubsection{Design principles.}
The Axon DSL was designed according to five main design principles.
\begin{enumerate}
\item\emph{Tensors and shapes are first-class citizens}:
Neural models and other components of state-of-the-art models resolve around the manipulation of tensors, i.e., potentially multi-dimensional arrays of values. We implemented tensors and their shapes as first-class citizens through a build-in tensor type and a dimension type, that acts both as symbolic dimension of a tensor shape in the type signature and as a referencable value in the executable part of the definition.
\item\emph{Expressive but intuitive syntax}:
Axon definitions should be expressive enough to be concise and intuitive enough to be readable. To this end, we adopted a Haskell-inspired syntax with do-style command sequences and pipe operators as intuitive surface representations of single-static assignment functional programs over tensors.
\item\emph{Clear denotational and operational semantics}:
A clear denotational and operational semantics allows to determine statically and dynamically what code means and how it will behave. We chose functional definitions as these rule out side effects and avoid hard-to-follow orthogonal and multi-layered abstractions common in many ML frameworks and backends.
\item\emph{Backend-agnostic primitive operations}:
The primitive operations should be backend-agnostic in the sense that most or all of them should be implementable for all the actual and potentially targetted backends. To this end, we avoided the use of primitive operations that only exist in one backend. The naming of the primitive operations does follow PyTorch naming in many instances simply because we assume this to be what most ML developers are at least somewhat acquainted with.
\item\emph{Balanced abstraction level}:
The abstract level should be balanced in the sense that modules remain relatively concise and high-level enough to understand but simultaneously are based on low-level primitive operations that can efficiently be implemented when lowering to backend code. We achieved this to mostly through refactoring similar code in reusable built-in libraries that hide the low-level operations from typical use of the high-level concepts such as attention or positional embeddings.
\end{enumerate}

\subsubsection{Implementation principles.}
The implementation of the Axon DSL follows three implementation principles.
\begin{enumerate}
\item\emph{Stable reusable IRs}:
All stages of the first phase of the compiler operate on the same canonical Axon AST, i.e., the representation of parsed, resolved, normalized, flattened, and typed programs differs only by invariants and metadata of the Axon AST rather than having different representations at each stage. Any Axon AST can be serialized as an Axon module.

All stages of the second phase operate on the same canonical Graph IR, i.e., rwrites and optimizations applied at the graph level operate within the semantic closure of the Graph IR. Any Graph IR can be serialized as an Axon module.

\item\emph{Stage-local contracts}:
Each stage owns a narrow contract, allowing validators and round-trip tests to target individual compiler properties. For example, name resolution should not infer tensor types. 

As an invariant, each stage should be stable under round-trips, i.e., serializing the results of a compiler stage, reparsing the resulting Axon module, and running it through the preceding stages should result in an identical Axon AST or Graph IR. This invariant is tested by serializing this IR and comparing it to the original serialization.

\item\emph{Limited model-specific code in shared modules}:
All operations needed to implement a model including highly model-specific quirks should be defined as Axon definitions in the respective Axon module. If an Axon definition can be reused by different modules, it can be promoted to a shared Axon module, either placed locally or as part of the built-in libraries.
\end{enumerate}

\subsection{Built-In Libraries}
\label{sec:axon-builtins}
Axon includes built-in libraries, which consist of Axon modules that cover the operations needed by most modern LLM architectures. Authors simply import reusable Axon definitions for standard layers, tensor manipulation, mathematics, masking, positions, caching, activations, state-space opertions, and mixture-of-expert logic. As these built-ins are themselves written in Axon, they are inspectable, type checked, and lowered through the same compiler pipeline as user-authored code. To be precise, the definitions form these built-in Axon modules are resolved and integrated into the main Axon module in the resolve stage such that later stages see a self-contained Axon module.

\paragraph{Standards layers, parameter, and configuration modules.}
The most central built-in modules are \lstinline{NN}, \lstinline{Params}, and \lstinline{Config}. \lstinline{NN} provides parameterized neural-network layers such as \lstinline{embedding}, \lstinline{linear}, \lstinline{layernorm}, and \lstinline{rmsnorm}. \lstinline{Params} exposes explicit checkpoint reads, while \lstinline{Config} provides typed access to configuration values. Listing~\ref{lst:axon-builtins-core} shows the typical authoring style.

\begin{lstlisting}[basicstyle=\ttfamily\scriptsize,caption={Core built-ins for layers, parameters, and configuration.},label={lst:axon-builtins-core}]
import Config
import NN
import Params (param)

embed_block@tok :: Tensor[B,S] -> Tensor[B,S,D]
embed_block ids = do
  hidden_dim <- Config.dim @@hidden_size
  x <- NN.embedding@tok ids dim=hidden_dim
  scale <- param @ln.weight
  x <- NN.rmsnorm@ln x
  return x * scale
\end{lstlisting}

Listing~\ref{lst:axon-builtins-core} illustrates the division of labor: configuration is read through \lstinline{Config}, checkpoint tensors through \lstinline{Params}, and backend-agnostic layers through \lstinline{NN}.

\paragraph{Tensor and math primitives.}
Lower-level tensor and scalar functionality lives in \lstinline{Tensor} and \lstinline{Math}. These modules provide reshaping, slicing, matrix multiplication, broadcasting, comparison, reductions, and elementary scalar functions. Listing~\ref{lst:axon-builtins-tensor} shows a small helper built directly from these primitives.

\begin{lstlisting}[caption={Tensor and math built-ins.},label={lst:axon-builtins-tensor},basicstyle=\ttfamily\scriptsize]
import Math (sqrt)
import Tensor

scale_heads :: Tensor[B,H,S,HD] -> Tensor[B,H,S,HD]
scale_heads x = do
  factor <- 1.0 / sqrt HD
  y <- Tensor.reshape x shape=[B, H, S, HD]
  return y * factor
\end{lstlisting}

Listing~\ref{lst:axon-builtins-tensor} is minimal, but the same modules also provide frequently used operators such as \lstinline{Tensor.transpose}, \lstinline{Tensor.concat}, \lstinline{Tensor.slice}, \lstinline{Tensor.softmax}, \lstinline{Tensor.where}, \lstinline{Math.exp}, and \lstinline{Math.clamp}.

\paragraph{Attention and masking.}
Transformer-specific definitions are packaged in \lstinline{Attention} and \lstinline{Masking}. These built-in modules cover head reshaping, grouped-QKV splitting, scaled dot-product attention, additive masking, grouped-query attention, and causal or bidirectional mask construction. Listing~\ref{lst:axon-builtins-attn} shows the standard self-attention pattern.

\begin{lstlisting}[caption={Attention and masking built-ins.},label={lst:axon-builtins-attn},basicstyle=\ttfamily\scriptsize]
import Attention
import Masking

self_attend :: Tensor[B,S,D] -> ?Tensor[B,K] -> Tensor[B,S,D]
self_attend x attn_mask = do
  q <- Attention.reshape_heads x heads=H
  k <- Attention.reshape_heads x heads=H
  v <- Attention.reshape_heads x heads=H
  keep <- Masking.causal_mask q k padding_mask=attn_mask
  y <- Attention.attention q k v keep
  return Attention.merge_heads y
\end{lstlisting}

Listing~\ref{lst:axon-builtins-attn} is representative of authored transformer code: model definitions express the architecture-level flow, while the builtin library packages the tensor mechanics behind typed interfaces.

\paragraph{Positions and cache management.}
Autoregressive decoding depends on two additional builtin families. \lstinline{Positions} computes position ids and rotary or relative-position ingredients, while \lstinline{Cache} manages null-safe key-value caching across decode steps. Listing~\ref{lst:axon-builtins-cache} shows the common preparation pattern.

\begin{lstlisting}[caption={Position and cache built-ins for decoding.},label={lst:axon-builtins-cache},basicstyle=\ttfamily\scriptsize]
import Cache
import Positions

decode_prep :: Tensor[B,S] -> ?Cache[B,H,T,C,DH] -> ?Tensor[B,K] -> ?Bool ->
  (Tensor[B,S], Tensor[B,S], ?Cache[B,H,T,C,DH])
decode_prep ids past attn_mask use_cache = do
  work_cache <- Cache.prepare past use_cache S
  past_len <- Cache.past_length past
  pos <- Positions.position_ids ids attn_mask past_length=past_len
  return ids, pos, work_cache
\end{lstlisting}

Listing~\ref{lst:axon-builtins-cache} captures the intended split: \lstinline{Positions} supplies index structure, \lstinline{Cache} threads state, and the backend only executes the lowered tensor program.

There are two \lstinline{Cache} built-in modules, the default one, which represents the the dynamic cache state as a growing list of pairs of tensors, and a \lstinline{static} one, which pre-allocates a statically-sized cache. Both \lstinline{Cache} built-in modules share the same interface, such that switching simply requires selecting the \lstinline{static} overlay. Depending on the target framework, this selection defaults to dynamic vs static for Pytorch vs JAX/MLX but can be overriden.

\paragraph{Activations and specialized modules.}
Beyond the core layers, Axon ships higher-level reusable libraries for common model families. \lstinline{Activations} provides functions such as \lstinline{gelu}, \lstinline{silu}, and \lstinline{swiglu}. \lstinline{MoE} provides routing and grouped expert helpers for mixture-of-experts models, and \lstinline{SSM} provides Mamba-style state-space building blocks. Listing~\ref{lst:axon-builtins-specialized} gives two representative calls.

\begin{lstlisting}[caption={Activation and specialized built-ins.},label={lst:axon-builtins-specialized},basicstyle=\ttfamily\scriptsize]
import Activations (silu)
import MoE

activate x = silu x

route_tokens@gate :: Tensor[B,S,D] -> (Tensor[B,S,TOPK], IdxTensor[B,S,TOPK])
route_tokens x =
  MoE.softmax_topk_router@gate x top_k=TOPK
\end{lstlisting}

Listing~\ref{lst:axon-builtins-specialized} shows the two ends of the spectrum: small reusable nonlinearities and larger architecture-specific helpers. The builtin boundary is therefore not limited to primitive operations; it also captures recurring structural patterns that should remain backend-independent.

Taken together, these modules form Axon's standard authoring surface. They keep model definitions concise and declarative, while still exposing enough structure for the compiler to typecheck, normalize, flatten, optimize, and lower the program through the shared AST and Graph IRs.

\subsubsection{Top-level declarations.}
An Axon module is a sequence of top-level items: definitions, global bindings, imports, exports, pragmas, and type aliases. Definitions may carry signatures, optional parameters, and scoped path parameters.

\begin{lstlisting}[caption={Top-level declarations, imports, exports, and a scoped definition.},label={lst:axon-ops-top-level},basicstyle=\ttfamily\scriptsize]
import NN
import Act (swiglu)
export block Pair

type Pair[T] = (T, T)

hidden_dim <- 768

block@attn :: Tensor[B,S,D] -> ?Tensor[B,K] -> Tensor[B,S,D]
block x attn_mask = do
  h <- NN.rmsnorm@ln x
  a <- Attention.attention h h h attn_mask
  return x + a
\end{lstlisting}

Listing~\ref{lst:axon-ops-top-level} groups the top-level forms that appear in ordinary authored Axon files. The first three lines show namespace import, member import, and export. The type alias introduces a reusable type constructor. The global binding defines a top-level constant expression. The definition itself combines a signature, an optional argument marked with \lstinline{?}, and a relative parameter scope \lstinline{@attn}.

\paragraph{Binding, return, and \texttt{do} blocks.}
The basic statement form is single-assignment binding with \lstinline{<-}. A \lstinline{do} block sequences binds and ends in an explicit \lstinline{return} or \lstinline{yield}.

\begin{lstlisting}[caption={Bindings and explicit return in a \texttt{do} block.},label={lst:axon-ops-do-block},basicstyle=\ttfamily\scriptsize]
pair_sum :: Tensor[B,D] -> Tensor[B,D] -> (Tensor[B,D], Tensor[B,D])
pair_sum x y = do
  s <- x + y
  d <- x - y
  return s, d
\end{lstlisting}

Listing~\ref{lst:axon-ops-do-block} shows the canonical block form. The same surface form also supports inline sequences separated by semicolons:

\begin{lstlisting}[caption={Inline \texttt{do} sequences.},label={lst:axon-ops-inline-do},basicstyle=\ttfamily\scriptsize]
identity x = do return x
cache_step x = do h <- NN.rmsnorm@ln x; return h
\end{lstlisting}

\paragraph{Conditionals.}
Axon provides both statement-level and expression-level conditionals, as well as ternary shorthand. Statement-level conditionals branch between two \lstinline{do} suites, while expression-level conditionals and ternaries produce values.

\begin{lstlisting}[caption={Expression-level and statement-level conditionals.},label={lst:axon-ops-conditionals},basicstyle=\ttfamily\scriptsize]
normalize_or_skip x use_norm =
  if use_norm then NN.rmsnorm@ln x else x

mask_or_null mask use_mask =
  use_mask ? mask : null

choose_path x use_skip = do
  if use_skip then do
    return x
  else do
    y <- NN.rmsnorm@ln x
    return y
\end{lstlisting}

Listing~\ref{lst:axon-ops-conditionals} places the three conditional forms side by side. Only the taken branch of a ternary or \lstinline{if/then/else} expression is evaluated. This is the one place where Axon is not eager.

\paragraph{Function application and composition.}
Calls are written in bare-argument style, optionally with keyword arguments. The language also exposes forward piping and monadic bind as first-class composition operators.

\begin{lstlisting}[caption={Bare calls, pipes, keyword arguments, and monadic bind.},label={lst:axon-ops-composition},basicstyle=\ttfamily\scriptsize]
mlp x = NN.linear@up x |> Act.swiglu |> NN.linear@down

mlp_kw x = NN.linear@proj x bias=true
mlp_bind x =
  NN.linear@up x >>= \h -> Act.swiglu h
\end{lstlisting}

Listing~\ref{lst:axon-ops-composition} shows the main composition idioms. Nested calls, pipes, lambda definitions, and binds are interchangeable surface styles for the same underlying dataflow. Pipes are often the clearest notation for feed-forward sublayers, while \lstinline{>>=} binds are convenient when an intermediate value must be named inside an expression.

\paragraph{Loops, carried state, and yielded values.}
The \lstinline{for} form iterates over explicit ranges. It supports optional lexical scoping via \lstinline{@...}, explicit steps, and carried values. A loop may appear as a statement or as a binding-producing form.

\begin{lstlisting}[caption={Ranges, carried values, and scoped \texttt{for} loops.},label={lst:axon-ops-loops},basicstyle=\ttfamily\scriptsize]
sum_to n = do
  total <- for i <- [0..n) carry (total) do
    next <- total + i
    yield next
  return total

states <- for@layers i <- [0..L) step=1 carry (cache) do
  layer <- Cache.index cache i
  yield layer
\end{lstlisting}

Listing~\ref{lst:axon-ops-loops} illustrates both looping forms. The first example updates a carried accumulator. The second shows the scoped form \lstinline{for@layers}, which binds the loop body under an additional lexical path prefix.

\paragraph{Scoped parameter binding.}
The \lstinline{scope} form introduces a lexical parameter path and binds the result of the nested sequence. This keeps reusable helper definitions independent of hard-coded checkpoint paths.

\begin{lstlisting}[caption={Scoped parameter binding with \texttt{scope}.},label={lst:axon-ops-scope},basicstyle=\ttfamily\scriptsize]
attn_out <- scope@attn do
  q <- NN.linear@q_proj x
  k <- NN.linear@k_proj x
  v <- NN.linear@v_proj x
  y <- Attention.attention q k v attn_mask
  return NN.linear@o_proj y
\end{lstlisting}

Listing~\ref{lst:axon-ops-scope} shows how a local lexical scope rewrites relative parameter paths. Inside the scope, relative paths such as \lstinline{@q_proj} or \lstinline{@o_proj} resolve against the enclosing \lstinline{@attn} prefix before lowering.

\paragraph{Values, paths, and type ascription.}
Expressions include names, literals, lists, tuples, parenthesized expressions, arithmetic, comparisons, Boolean connectives, and explicit type ascriptions.

\begin{lstlisting}[caption={Values, paths, comparisons, and type ascription.},label={lst:axon-ops-values},basicstyle=\ttfamily\scriptsize]
spec ids = do
  dims <- [B, S, D]
  pair <- (ids, ids)
  abs_path <- @@model.embed_tokens
  rel_path <- @mlp.c_fc
  same <- ids == ids
  typed <- (ids :: Tensor[B,S])
  return dims, pair, abs_path, rel_path, same, typed
\end{lstlisting}

Listing~\ref{lst:axon-ops-values} collects the basic value forms. At the type level, Axon supports optional types, tuple types, list types, tensor types with symbolic dimensions, and symbolic-dimension arithmetic such as \lstinline{Tensor[B,S,H+Hd]}. Together these forms cover the full surface language accepted by the grammar while keeping authored model definitions compact and auditable.

\paragraph{Extended Backus-Naur Grammar}
We provide the Backus-Naur form in Figure \ref{lst:axon-grammar}, for Axon’s surface syntax as a readable specification of the language accepted by the compiler.

\subsection{Axon Compiler Phases}\label{sec:appendix:phases}

The Axon compiler comprises three main phases: (1) desugaring of full Axon DSL into its semantic core; (2)  optimization at the Graph IR level; and (3) lowering of the Graph IR to backend-specific code.
In the technical supplement, we walk an exemplary small definition through all phases.

\paragraph{Phase 1: Desugaring.}
The Axon DSL is defined by an EBNF grammar that is operationalized using Lark \cite{lark}. The parser transforms valid Axon DSL into an abstract syntax tree (Axon AST), which serves as the representation for Phase 1 and can be dumped to Axon DSL and reparsed. Every desugaring stage of Phase 1 receives an Axon AST and outputs an Axon AST, typically guaranteeing additional properties: (a) the resolver stage inlines library definitions imported and referenced, guaranteeing a closure property stating that all non-primitive operations are expressed as definitions in the AST; (b) the flattening stage flattens nested expressions, guaranteeing a flatness property stating that definitions consist of a flat sequence of operations; (c) the type checker stage infers types from type signatures, type ascriptions, and the type rules of primitive operations, guaranteeing a consistently typed Axon AST. 
The desugaring phase also contains dead code elimination, constant folding, and other straightforward code normalizations.

\paragraph{Phase 2: Graph optimizations.}
The closed flat typed Axon AST can be viewed as an instance of a static single assignment (SSA) graph. In this phase, the Axon compiler converts this semantic core to the Graph IR, which is then iteratively optimized using a number of stages: \emph{(a) definition inlining} replaces wrapper definitions and other low-complexity definitions by integrating their sequence of operations into their call sites, enabling, for example, the inlining of primitive wrapper definitions; \emph{(b) generic graph rewriting} replaces known patterns of operations by equivalent patterns of operations based on matching of primitive operational provenance, enabling, for example, the replacement of loops iterating over experts in an MoE by a more efficient stacked-experts formulation; and \emph{(c) backend-specific graph rewriting} replaces known patterns of operations by backend-specific primitives, enabling, for example, the use of optimized attention implementations and kernels.

\paragraph{Phase 3: Lowering.}
The optimized Graph IR contains definitions as assignment sequences, calls to definitions, and generic and backend-specific primitive operations. The lowering phase is backend-specific and transpiles the Graph IR into one of the currently targeted backends: PyTorch, Triton-accelerated PyTorch, MLX, JAX, and vLLM. Each of the separate code generators needs to be able to implement the control flow and the primitive operators represented by the Graph IR. In addition, the code generators also need to implement \texttt{forward} and \texttt{generate} methods. To achieve competitive performance, compilation-oriented backends typically supply scaffolding that ensures static cache allocations.

\subsection{Axon Compiler Stages}

Table~\ref{tab:phase-invariants} lists each stage of the three phases of the Axon compiler together with the representation it operates on and the invariant it establishes. As is conventional, the first invariant is monotonicity of structure: later compiler stages should remove ambiguity and add metadata but should not throw away semantic information or re-interpret the program. The second invariant is backend isolation: backend implementations may choose different tensor libraries and primitive implementations, but they must consume the same Graph IR contract. If a backend requires weaker validation or model-family-specific dispatch to run, the issue belongs within the backend implementation.

\begin{table*}[t] 
\centering
\caption{Axon Compiler Phase and Stage with Invariants}
\label{tab:phase-invariants}
\begin{tabularx}{\textwidth}{llX} 
\toprule
Phase & Representation & Primary invariant \\
\midrule
Parse & one-file AST & Syntactic structure and explicit MAIN pragma insertion \\
Load & loaded AST set & Imports and builtins located without rewriting semantics \\
Materialize & one-file AST & Optional checkpoint/config specialization for generic models \\
Resolve/validate-closed & closed AST & No unresolved imports or names; unreachable definitions pruned from MAIN \\
Normalize & normalized AST & Call syntax, pipes, path sugar, and zero-arg call/name distinctions made explicit \\
Elaborate/validate-elaborated & elaborated AST & Default arguments filled and call arguments positionalized \\
Flatten/validate-flat & flat AST & Explicit evaluation order; flat calls and binds accepted by typecheck and Graph IR lowering \\
Typecheck/validate-typed & Typed flat AST & expression types, arities, dimensions, and primitive rules applied to a fixpoint \\
Optimize-ast & typed flat AST & Optional conservative AST cleanup with retype/validation \\
Graph lowering/validation & Graph IR & Typed graph modules, multi-output nodes, structured paths, constraints, and metadata \\
Optimize-graph/validation & Graph IR & Optional graph cleanup, specialization, backend-neutral rewrites, and opt-in backend intrinsics \\
Backend & generated/runtime code & Executable tensor program consuming the validated Graph IR contract \\
\bottomrule
\end{tabularx}
\end{table*}

\begin{table*}[t]
\centering
\caption{Wall-clock time and GPU idle for autoregressive generation (112 tokens for Qwen 0.5B, 96 for Pleias 3.8B).
``---'' indicates the metric is not directly measurable: JAX has no \lstinline{torch.profiler} equivalent for per-kernel timing, and GPU active/idle is a synchronous-execution concept that does not apply to JAX's async dispatch model.}

\label{tab:gpu}
\resizebox{\textwidth}{!}{
\begin{tabular}{lcccccc}
\toprule
& \multicolumn{3}{c}{Qwen2.5-0.5B (0.67B)} & \multicolumn{3}{c}{Pleias-3b-Preview (3.8B)} \\
\cmidrule(lr){2-4} \cmidrule(lr){5-7}
Metric & Transformers & Axon-Torch & Axon-JAX & Transformers & Axon-Torch & Axon-JAX \\
\midrule
Wall-clock (ms) & 1380 & 1185 & \textbf{541} & 1091 & \textbf{1071} & 1160 \\
Speedup vs Transformers & 1.0$\times$ & 0.86$\times$ & \textbf{0.39$\times$} & 1.0$\times$ & 0.98$\times$ & 1.07$\times$ \\
GPU active (ms) & 373 & 370 & --- & 557 & 561 & --- \\
GPU idle (ms) & 1007 & 815 & --- & 534 & 510 & --- \\
GPU idle \% & 73.0\% & 68.8\% & 0.0\% & 48.9\% & 47.6\% & 0.0\% \\
CUDA kernels & 129{,}189 & 120{,}987 & --- & 101{,}935 & 104{,}123 & --- \\
\bottomrule
\end{tabular}
}
\end{table*}

\begin{table*}[t]
\centering
\caption{Python function call counts and cProfile cumulative time.
``---'' indicates the category does not apply: JAX fuses all per-op dispatches into a single XLA compilation per step, so individual \lstinline{F.linear} and \lstinline{SDPA} calls do not exist as Python-level dispatches.}
\label{tab:calls}
\resizebox{\textwidth}{!}{
\begin{tabular}{lcccccc}
\toprule
& \multicolumn{3}{c}{Qwen2.5-0.5B (0.67B)} & \multicolumn{3}{c}{Pleias-3b-Preview (3.8B)} \\
\cmidrule(lr){2-4} \cmidrule(lr){5-7}
Metric & Transformers & Axon-Torch & Axon-JAX & Transformers & Axon-Torch & Axon-JAX \\
\midrule
Total Python calls & 455{,}769 & 508{,}390 & \textbf{137{,}805} & 359{,}909 & 330{,}929 & \textbf{82{,}210} \\
Calls vs Transformers & 1.0$\times$ & 1.12$\times$ & 0.30$\times$ & 1.0$\times$ & 0.92$\times$ & 0.23$\times$ \\
cProfile time (s) & 1.13 & 0.95 & \textbf{0.54} & 0.87 & \textbf{0.78} & 1.16 \\
\midrule
\multicolumn{7}{l}{\emph{Per-op dispatch (Python calls per generate pass):}} \\
\quad \lstinline{nn.Module.__call__} & 35{,}616 & 0 & 0 & 28{,}032 & 0 & 0 \\
\quad \lstinline{F.linear} & 18{,}928 & 18{,}928 & --- & 14{,}880 & 14{,}880 & --- \\
\quad \lstinline{SDPA} & 2{,}688 & 2{,}688 & --- & 2{,}112 & 2{,}112 & --- \\
\quad \lstinline{rope_apply} & 0 & 5{,}376 & --- & 0 & 4{,}224 & --- \\
\quad \lstinline{forward} (jit dispatch) & --- & --- & 112 & --- & --- & 96 \\
\bottomrule
\end{tabular}
}
\end{table*}

\subsection{Running Example}\label{sec:appendix:running-example}

In this section we walk a small Axon fragment through the main compiler stages. The example is intentionally smaller than a full transformer block, but exercises the same mechanisms: path sugar, defaulted keyword arguments, tensor shapes, nested calls, and typed graph lowering.

\paragraph{Surface Axon.}
A reusable projection can be written in a concise functional form:
\begin{lstlisting}[basicstyle=\ttfamily\scriptsize]
project :: Tensor[B,S,D] -> Tensor[B,S,O]
project x = do
  y <- Tensor.reshape x shape=[B, S, D]
       |> NN.linear@proj dim=O
  return y
\end{lstlisting}
This surface program contains a pipe, callee-attached path sugar \lstinline{@proj}, a shape expression, and omitted default arguments such as \lstinline{bias=true} from the library definition of \lstinline{NN.linear}.

\paragraph{Normalize and elaborate.}
Normalization removes the pipe syntax and path sugar by turning them into ordinary call structure. Elaboration then fills in the default parameters declared by the callee signature.
\begin{lstlisting}[basicstyle=\ttfamily\scriptsize]
project x = do
  y <- NN.linear @proj (Tensor.reshape x shape=[B, S, D]) dim=O bias=true
  return y
\end{lstlisting}
The point is not that all defaults must be textually present in the surface program. The invariant is that downstream stages see a call surface whose positional, keyword, and path
arguments are unambiguous.

\paragraph{Flatten.}
Flattening preserves eager evaluation order by introducing explicit binds and making paths absolute or templated absolute values:
\begin{lstlisting}[basicstyle=\ttfamily\scriptsize]
project :: Tensor[B,S,D] -> Tensor[B,S,O]
project x = do
  _v1 <- Tensor.reshape x shape=[B, S, D]
  y <- NN.linear @@proj _v1 dim=O bias=true
  return y
\end{lstlisting}
The flat program has no pipe expression, no nested call used as an argument, and no callee path sugar. This is the Axon AST form expected by the typechecker and trivally translated into the Graph IR.

\paragraph{Typecheck.} Typecheck annotates every expression with type, arity, and tensor-dimension metadata. The same flat program can be read as: 
\begin{lstlisting}[basicstyle=\ttfamily\scriptsize]
_v1 <- (Tensor.reshape (x :: Tensor[B,S,D])
          shape=([B, S, D] :: List[Dim]) :: Tensor[B,S,D])
y <- (NN.linear (@@proj :: Path) (_v1 :: Tensor[B,S,D])
        dim=(O :: Dim) bias=(true :: Bool) :: Tensor[B,S,O])
\end{lstlisting}
What matters is not the rendered punctuation, but that the shape relation between \lstinline{Tensor[B,S,D]} and \lstinline{Tensor[B,S,O]} is explicit before lowering. 

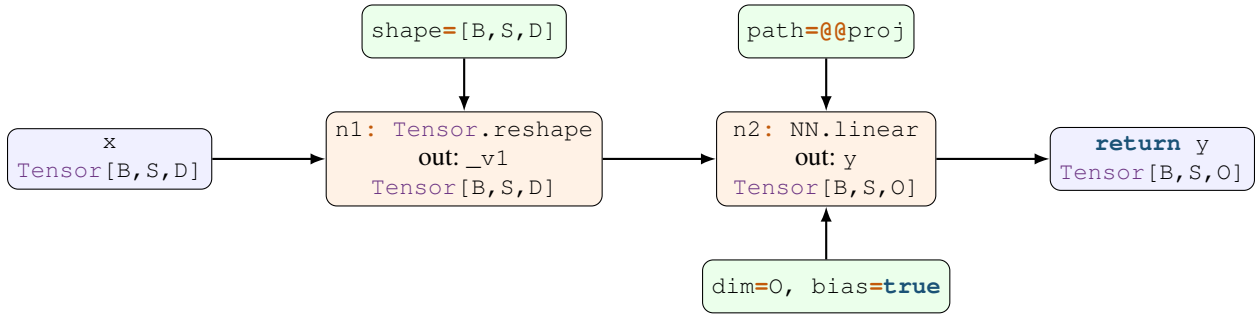
\begin{figure*}[t]
\centering
\begin{tikzpicture}[
  value/.style={draw, rounded corners, fill=blue!6, align=center, minimum width=2.3cm, minimum height=0.8cm},
  nodeop/.style={draw, rounded corners, fill=orange!10, align=center, minimum width=2.9cm, minimum height=1.0cm},
  attr/.style={draw, rounded corners, fill=green!8, align=center, minimum width=2.2cm, minimum height=0.7cm},
  edge/.style={-{Latex[length=2mm]}, thick}
]
\node[value] (x) {\lstinline{x}\\\lstinline{Tensor[B,S,D]}};
\node[nodeop, right=1.5cm of x] (reshape) {\lstinline{n1: Tensor.reshape}\\out: \lstinline|_v1|\\\lstinline{Tensor[B,S,D]}};
\node[nodeop, right=1.5cm of reshape] (linear) {\lstinline{n2: NN.linear}\\out: \lstinline{y}\\\lstinline{Tensor[B,S,O]}};
\node[value, right=1.5cm of linear] (ret) {\lstinline{return y}\\\lstinline{Tensor[B,S,O]}};
\node[attr, above=0.7cm of reshape] (shape) {\lstinline{shape=[B,S,D]}};
\node[attr, above=0.7cm of linear] (path) {\lstinline{path=@@proj}};
\node[attr, below=0.7cm of linear] (attrs) {\lstinline{dim=O, bias=true}};
\draw[edge] (x) -- (reshape);
\draw[edge] (reshape) -- (linear);
\draw[edge] (linear) -- (ret);
\draw[edge] (shape) -- (reshape);
\draw[edge] (path) -- (linear);
\draw[edge] (attrs) -- (linear);
\end{tikzpicture}
\caption{Graph IR view of the running example. Blue denotes typed graph values, orange
operation nodes, and green structured operands and attributes. Paths, literals, and shape
lists are structured operands, not strings that backends must parse.}
\label{fig:running-graph-ir}
\end{figure*}

\paragraph{Graph IR.}
Lowering turns each flat bind into a typed graph node. A representation of the graph module for \lstinline{project} is given in Figure~\ref{fig:running-graph-ir}, which might be schematically represented as follows:
\begin{lstlisting}[basicstyle=\ttfamily\scriptsize]
module project(x: Tensor[B,S,D]) -> Tensor[B,S,O]
  n1 = call Tensor.reshape(
         x, shape=[B,S,D]) -> Tensor[B,S,D]
  n2 = call NN.linear(
         path=@@proj, x=n1, dim=O, bias=true)
       -> Tensor[B,S,O]
  return n2
\end{lstlisting}

\paragraph{Graph Optimizations.}
As already introduced in the description of Phase 2, the graph optimizations rewrite, replace, or join module graphs at the Graph IR level. An exhaustive list would be beyond the scope of this appendix.

\paragraph{Backend.}
The Torch, Triton, MLX, JAX, and vLLM code generators and their runtime implementations are based on the Graph IR and therefore share the same typed generic primitive operations, path values, and output types, which defines the cross-backend interface at the lower levels of the Axon compiler pipeline. The only differences are backend-specific primitive operations.

\subsection{A Complete Model Definition}\label{sec:appendix:complete-model} 

Figure~\ref{lst:gpt2} gives the complete Axon module GPT-2. The entire model, including the layer stack, is expressed in these Axon definitions and the Axon definitions of the built-in modules uses. The PyTorch, Triton, JAX, MLX, and vLLM implementations benchmarked in this paper are all generated from these Axon definitions together with a standard \texttt{safetensors} or \texttt{PyTorch} checkpoint.

\begin{figure*}[t]
    \caption{Complete Axon definition of GPT-2.}
    \label{lst:gpt2}
    \begin{lstlisting}
{-# CHECKPOINTS ["openai-community/gpt2", "openai-community/gpt2-medium", "openai-community/gpt2-large", "openai-community/gpt2-xl"] #-}
{-# TASK "causal_lm" #-}

import Activations (gelu_new)
import Attention (attention, merge_heads, reshape_heads)
import Masking (Mask, causal_mask_for_input, mask_for_input, mask_length_for_input)
import Positions (position_ids)
import Cache
import Config
import NN
import Tensor

CONTEXT_SIZE <- (Config.dim @@n_positions default=1024 :: Dim)
MODEL_DIM <- (Config.dim @@n_embd default=768 :: Dim)
NUM_LAYERS <- (Config.dim @@n_layer default=12 :: Dim)
NUM_HEADS <- (Config.dim @@n_head default=12 :: Dim)
VOCAB_SIZE <- (Config.dim @@vocab_size default=52057 :: Dim)

gpt2_block :: Tensor[B,S,D] -> Tensor[B,1,S,K] -> ?CacheLayer[B,H,P,CONTEXT_SIZE,DH] -> (Tensor[B,S,D], ?CacheLayer[B,H,K,CONTEXT_SIZE,DH])
gpt2_block x mask past_kv = do
  x1 <- NN.layernorm@ln_1 x
  a, new_kv <- scope@attn do
    q_lin, k_lin, v_lin <- Tensor.chunk (NN.linear@c_attn x1 dim=(3 * MODEL_DIM) bias=true transpose=true) parts=3
    q <- reshape_heads q_lin heads=NUM_HEADS
    k <- reshape_heads k_lin heads=NUM_HEADS
    v <- reshape_heads v_lin heads=NUM_HEADS
    k, v, new_kv <- Cache.update past_kv k v
    a <- attention q k v mask |> merge_heads |> NN.linear@c_proj dim=MODEL_DIM bias=true transpose=true
    return a, new_kv
  x <- x + a
  x3 <- NN.layernorm@ln_2 x
  m <- scope@mlp do
    return NN.linear@c_fc x3 dim=(4 * MODEL_DIM) bias=true transpose=true |> gelu_new |> NN.linear@c_proj dim=MODEL_DIM bias=true transpose=true
  return x + m, new_kv

gpt2 :: Tensor.TokenIds[B,S] -> ?Mask[B,K,CONTEXT_SIZE] -> ?Cache[B,NUM_HEADS,P,CONTEXT_SIZE,MODEL_DIM / NUM_HEADS] -> ?Bool -> (Tensor[B,S,V], ?Cache[B,NUM_HEADS,K,CONTEXT_SIZE,MODEL_DIM / NUM_HEADS])
gpt2 input_ids attn_mask past_kv use_cache = do
  tok <- NN.embedding@wte input_ids dim=MODEL_DIM
  past_len <- Cache.past_length past_kv
  attn_mask <- mask_for_input input_ids attn_mask CONTEXT_SIZE
  pos <- position_ids input_ids attn_mask=attn_mask past_length=past_len pad_fill=1 |> NN.embedding@wpe
  x <- tok + pos
  key_len <- mask_length_for_input input_ids attn_mask past_len
  mask <- causal_mask_for_input input_ids key_len padding_mask=attn_mask window=CONTEXT_SIZE
  work_kv <- Cache.prepare past_kv use_cache CONTEXT_SIZE
  read_kv <- Cache.read past_kv work_kv
  x, work_kv <- for@h i <- [0..NUM_LAYERS) carry (x, work_kv) do
    past_i <- Cache.index read_kv i
    x, new_i <- gpt2_block x mask past_i
    work_kv <- Cache.append work_kv new_i
    yield x, work_kv
  new_length <- past_len + S
  work_kv <- Cache.finish work_kv new_length
  new_kv <- (use_cache ? work_kv : null)
  logits <- NN.layernorm@ln_f x |> NN.linear@wte
  return logits, new_kv

    \end{lstlisting}
\end{figure*}

\subsection{Python Overhead: Axon vs. Transformers}
\label{sec:appendix:python-overhead}
We profile autoregressive generation (max. 128 tokens) on two models: Qwen/Qwen2.5-0.5B
(0.67B parameters, 24 layers) and PleIAs/Pleias-3b-Preview (3.8B, 22 layers) to investigate the Python dispatch overhead that Axon's codegen eliminates. Both
models were selected because they achieved $>1.0\times$ speedup on both the PyTorch and JAX backends in a 91-checkpoint benchmark, naturally representing subjects of interest for further analysis.

All configurations run on the same underlying framework. Axon's \lstinline{torch} code generator emits standard PyTorch code that calls the same \lstinline{torch._C._nn.linear}, \lstinline{scaled_dot_product_attention}, and CUDA kernels as HuggingFace's Transformers. Axon's \lstinline{jax} code generator emits standard JAX code compiled by XLA. The speedup therefore cannot come from \emph{better} kernels. Hence, it comes from how the code is structured at the Python layer.

Table~\ref{tab:gpu} reports wall-clock time and GPU idle percentage for each configuration.  GPU active time is measured via \lstinline{torch.profiler} (CPU+CUDA activities), which sums \lstinline{self_device_time_total} across all CUDA kernel events.  GPU idle is wall-clock time minus GPU active time.

For the JAX backend, GPU idle is reported as 0.0\%.  This is a property of JAX's asynchronous dispatch model. JAX enqueues operations on the accelerator without blocking Python, returning \lstinline{jax.Array} futures\footnote{https://docs.jax.dev/en/latest/async\_dispatch.html}. Python code runs ahead of the GPU, so the device always has queued work and never idles waiting for dispatch. This is not a measurement artifact, JAX simply has no concept of GPU idle between kernel launches because there is no synchronous gap between them.  We verified that wall-clock measurements use \lstinline{jnp.block_until_ready()} to ensure the GPU has completed all queued work before recording the end time.

Table~\ref{tab:calls} reports total Python function calls and cProfile cumulative time.  Python call counts are measured by \lstinline{cProfile} in a separate profiling pass (after the wall-clock pass) to avoid profiler overhead interference.  JAX's \lstinline{cProfile} time on Pleias 3.8B (1.16\,s) exceeds Tranformer's (0.87\,s) despite having 77\% fewer calls. The remaining calls include expensive JAX runtime primitives (\lstinline{apply_primitive} for \lstinline{broadcast}/\lstinline{full} operations) that dominate this model size.

The comparison is fair by construction: Axon's \lstinline{torch} code generator emits PyTorch code that calls the \emph{same} \lstinline{torch._C._nn.linear}, \lstinline{scaled_dot_product_attention}, and CUDA kernels as HuggingFace's Transformers.  Both produce identical token sequences ($100\%$ top-1 parity). The speedup comes entirely from the Python layer \emph{above} the kernels.

\paragraph{The Price of Transformers.}
Transformers implements each model as a nested hierarchy of \lstinline{nn.Module} subclasses. Every \lstinline{__call__} goes through PyTorch's dispatch chain (\lstinline{__call__} $\to$ \lstinline{_wrapped_call_impl} $\to$
\lstinline{_call_impl} $\to$ \lstinline{forward}), costing 3 Python function calls per module per step per layer. For Qwen2.5-0.5B (24 layers, $\sim$13 modules/layer), this is \textbf{35{,}616 wrapper calls} per generate pass. On top of this, Transformers's \lstinline{generate()} loop adds per-step \lstinline{prepare_inputs_for_generation}, a \lstinline{LogitsProcessorList} chain (float32 cast + clone), and a stopping-criteria check.

\paragraph{Advantages of Axon.}
Axon's compiler pipeline (resolve $\to$ normalize $\to$ elaborate $\to$ typecheck $\to$ lower $\to$ graph optimize $\to$ emit) flattens the module hierarchy into a single inline function: \lstinline{model.forward()} $\to$ \lstinline{_def_qwen2()} (one flat function, zero module wrappers).

Weight access changes from attribute chains through \lstinline{nn.Module.__getattr__} to direct dict lookups (\lstinline{_param("layers.0.self_attn.q_proj.weight")}).  The generate loop
drops from 6 Python phases to 4 (no logits processor, no \lstinline{prepare_inputs}).  Table~\ref{tab:calls} confirms: the GPU-op call counts (\lstinline{F.linear}, \lstinline{SDPA}) are \emph{identical} between Transformers and Axon-Torch, but \lstinline{nn.Module.__call__} drops from 35{,}616 to 0.

\paragraph{Comparison to JAX.}
Axon's \lstinline{jax} code generator applies the same flattening but \lstinline{jax.jit} fuses all per-op Python dispatches into a single XLA compilation per step.  The 18{,}928 \lstinline{F.linear} calls and 2{,}688 \lstinline{SDPA} calls that exist as separate Python dispatches in both Transformers and Axon-Torch become \textbf{zero} Python calls, they are compiled into one fused XLA kernel.  Combined with JAX's async dispatch, which lets Python run ahead of the GPU (0\% GPU idle), this achieves 2.55$\times$ speedup on the 0.5B model. However, the per-step host synchronization for \texttt{EOS} checking (\lstinline{jax.device_get}) scales with GPU compute time, causing JAX's advantage to diminish on larger models.

\paragraph{Why this generalizes.}
The overhead Transformers Library pays is structural, not model-specific:
\lstinline{nn.Module.__call__} dispatch scales with $\text{num\_layers} \times \text{modules\_per\_layer}$. Logits processing happens every step regardless of model. \lstinline{prepare_inputs_for_generation} is a Python abstraction every model passes through. Axon's codegen eliminates these for \emph{any} model. The compiler flattens the hierarchy at compile time, so the generated code never pays the \lstinline{nn.Module} dispatch tax.  This is why the speedup is consistent across architectures (Qwen, Mistral, Llama, Bloom, Pleias) and scales with decode steps (generate is faster; forward is not, because a single pass amortizes the overhead).

\subsection{Full-width versions of the main-text figures.}
Figures~\ref{fig:app:bench4b:autoreg}--\ref{fig:app:mlx} reproduce the remaining benchmark
figures from the main text at full page width.
 
\begin{figure*}[p]
  \centering
  \includesvg[width=\textwidth]{Figures/4b-bench/scatter_all_gen.svg}
  \caption{Autoregressive generation performance with decoder-only models with up to $4B$ parameters. Log-log plot where the diagonal line indicates equal performance. Points above the parity line indicate Axon is faster, color-fill denotes backends and border-line denotes dtype.}
  \label{fig:app:bench4b:autoreg}
\end{figure*}
 
\begin{figure*}[p]
  \centering
  \includesvg[width=\textwidth]{Figures/32b-bench/scatter_all_gen_32b.svg}
  \caption{Benchmarking Autoregressive models between $4B$ and $32B$ models. 
  Log-log plot where the diagonal parity line indicates equal performance. Generally, Axon produces comparable or superior throughput on the bigger models.}
  \label{fig:app:bench32b:autoreg}
\end{figure*}
 
\begin{figure*}[p]
  \centering
  \includesvg[width=\textwidth]{Figures/vllm-bench/scatter_vllm_native_gen_all.svg}
  \caption{vLLM: Axon (vLLM native) vs. Transformers generation throughput on 88 checkpoints. Points above the parity line indicate Axon is faster. $74\%$ fall above.}
  \label{fig:app:vllm}
\end{figure*}
 
\begin{figure*}[p]
  \centering
  \includesvg[width=\textwidth]{Figures/mlx-bench/twocol-tok-s-scatter-len256.svg}
  \caption{MLX. Benchmarking conventional Transformers models against Axon derived standalone model definitions. Axon yields some considerable speed-ups across the board.}
  \label{fig:app:mlx}
\end{figure*}

\begin{figure*}[p]
  \centering
  \includesvg[width=\textwidth]{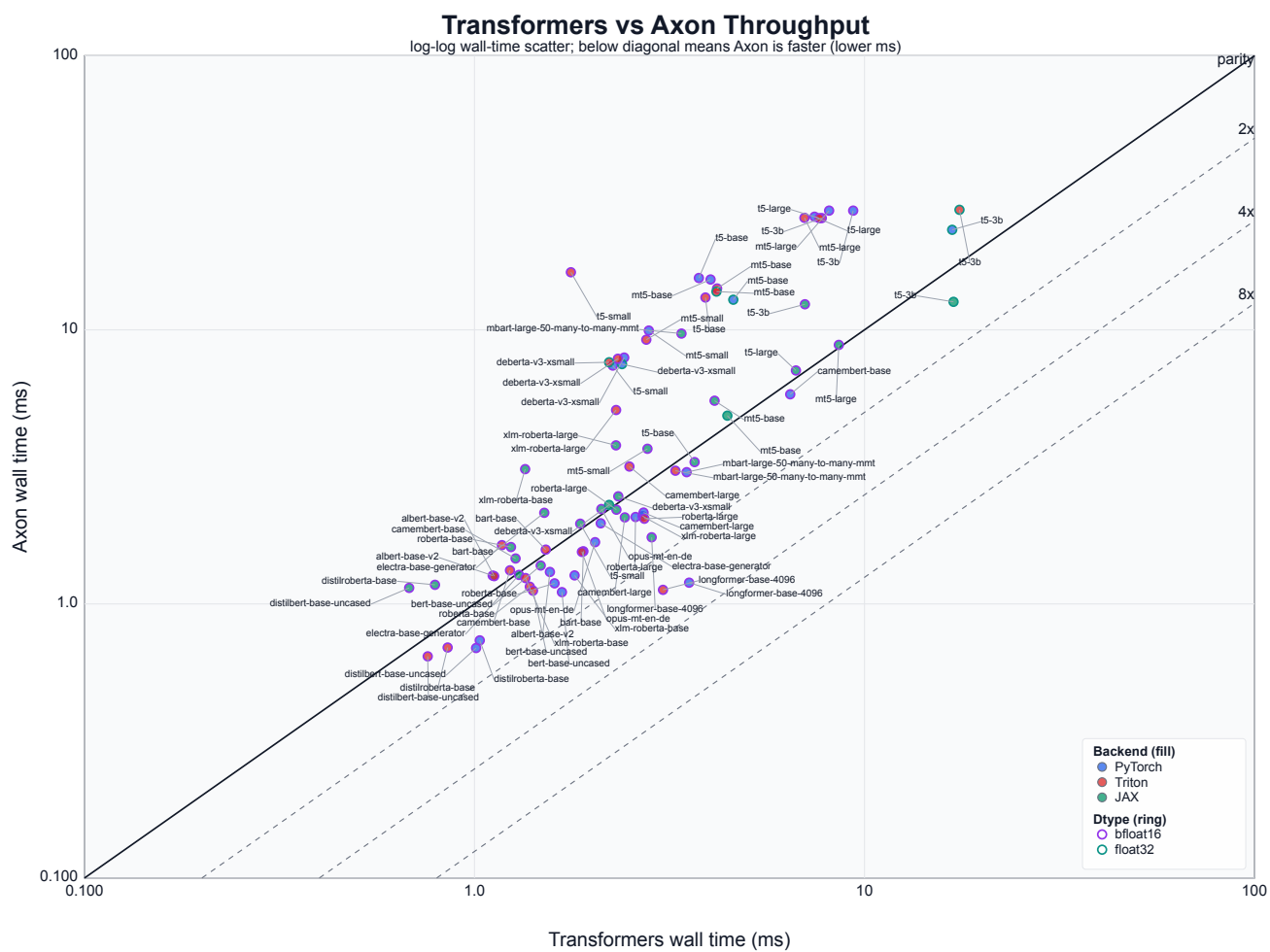}
  \caption{Benchmarking Encoder-only and Encoder--Decoder up to $4B$ models. Log-log plot where the diagonal parity line indicates equal performance. Here, we see the majority of the points in the vicinity below the parity line.}
  \label{fig:bench4b:enc-and-enc-dec}
\end{figure*}

\begin{figure*}[p]
  \centering
  \includesvg[width=\textwidth]{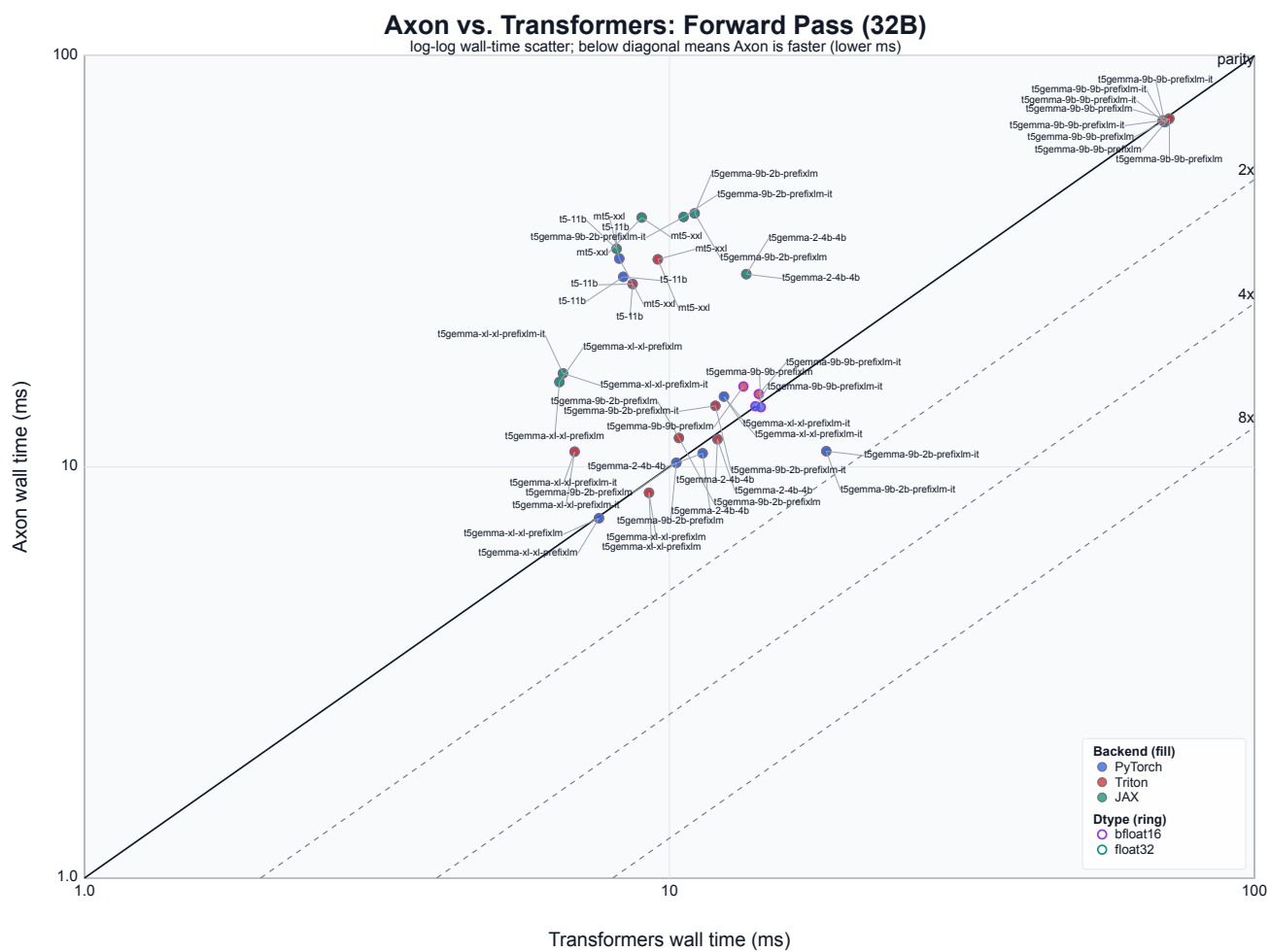}
    \caption{Benchmarking Encoder-only and Encoder-Decoder between $4B$ and $32B$ models. Log-log plot where the diagonal parity line indicates equal performance.}
  \label{fig:bench32b:enc-and-enc-dec}
\end{figure*}

\paragraph{Encoder-only and encoder--decoder models.}
Figure~\ref{fig:bench4b:enc-and-enc-dec} shows the performance comparison for encoder-only and encoder-decoder models up to $4B$.
Figure~\ref{fig:bench32b:enc-and-enc-dec} shows the performance comparison for encoder-only and encoder-decoder models between $4B$ and $32B$.

\begin{figure*}[p]
  \centering
  \includesvg[width=\textwidth]{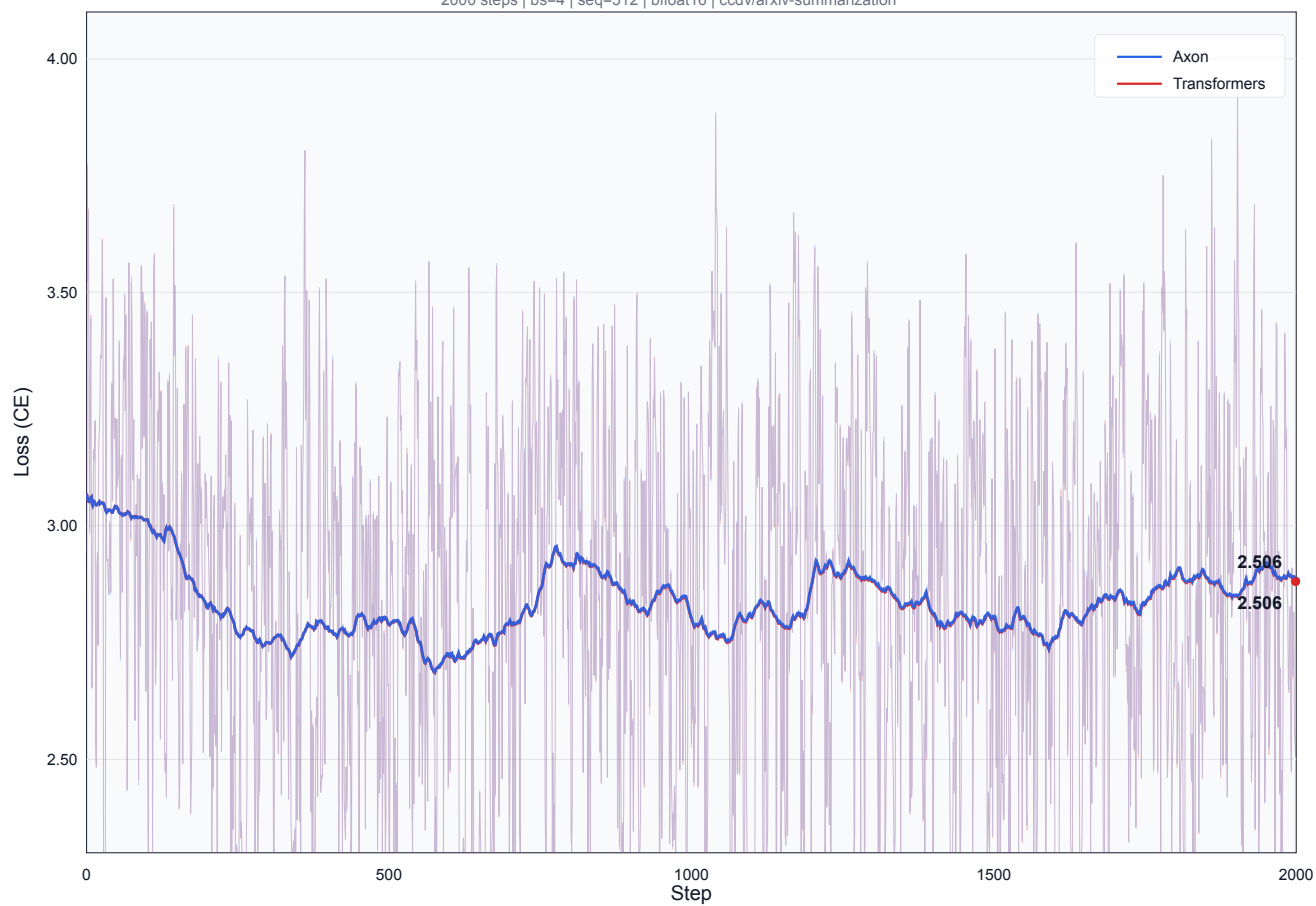}
  \caption{Training of Gemma 3 270M on summarization task for 2000 steps, demonstrating identical training behaviour between conventional Transformers-definition and Axon derived PyTorch model. Note the Axon and Transformers loss curves are on top of each other together.}
  \label{fig:axon-training}
\end{figure*}
Figure~\ref{fig:axon-training} shows a loss curve comparison on Gemma 3 270M between Transformer's implementation and an Axon-derived PyTorch model. The important insight here is not the quality of the training but the identical training behaviour with the two curves perfectly overlaid on each other.

\onecolumn
\begin{lstlisting}[
    caption={Axon Grammar.},
    label={lst:axon-grammar},
    basicstyle=\ttfamily\small,
    breaklines=true,
    frame=single,
]
# Axon Grammar (Current Authoring Form)
#
# Indentation-sensitive (Python-style INDENT/DEDENT tokens).
# "--" starts a comment outside strings.
# Newlines are significant; _NL tokens carry indentation context.
# The NAME terminal is permissive: it matches qualified identifiers
# including "::", "@", and "." internally (e.g. ns::op, mod@path, ns.mod).
#
# Authoritative source: brainsurgery/synapse/axon/parse/grammar.py
# (Lark LALR grammar with custom indentation post-lexer).
#
# Some parser-only constructs (parallel arg_expr hierarchy, nl_gap/arg_ws
# whitespace gaps for multi-line pipe/ternary/if expressions) are omitted
# for clarity; see grammar.py for the full Lark grammar.

program             ::= { top_item NEWLINE } ;

top_item            ::= definition_decl
                      | global_binding
                      | import_decl
                      | export_decl
                      | pragma
                      | type_alias_decl
                      ;

# -- Module definitions --

definition_decl     ::= [ signature NEWLINE ] definition ;
signature           ::= mod_decl "::" signature_type ;
signature_type      ::= type_expr { "->" type_expr } ;
definition          ::= mod_decl { def_param } "=" expr ;
mod_decl            ::= module_name { "@" NAME } ;
module_name         ::= NAME { "." NAME } ;
def_param           ::= NAME
                      | "?" NAME "=" def_param_simple
                      | "?" NAME "=" "(" expr ")"
                      ;
def_param_simple    ::= literal | NAME ;

# -- Global bindings --

global_binding      ::= module_name "<-" expr ;

# -- Imports --

import_decl         ::= "import" import_item { "," import_item } ;
import_item         ::= module_name [ import_members ] ;
import_members      ::= "(" [ NAME { "," NAME } [ "," ] ] ")"
                      | NAME { NAME }
                      ;

# -- Exports --

export_decl         ::= "export" export_members ;
export_members      ::= "(" [ NAME { "," NAME } [ "," ] ] ")"
                      | NAME { NAME }
                      ;

# -- Pragmas --

pragma              ::= "{-#" NAME pragma_value "#-}" ;
pragma_value        ::= tuple_expr | list_expr | literal ;

# -- Type aliases --

type_alias_decl     ::= "type" NAME [ type_alias_params ] "=" type_expr ;
type_alias_params   ::= "[" type_alias_param { "," type_alias_param } "]" ;
type_alias_param    ::= NAME | ".." NAME ;

# -- Type expressions --

type_expr           ::= type_optional ;
type_optional       ::= "?" type_optional
                      | type_tuple
                      | type_list
                      | type_tensor
                      | type_name
                      ;
type_tuple          ::= "(" type_expr "," type_expr { "," type_expr } [ "," ] ")" ;
type_list           ::= "List" "[" type_expr "]" ;
type_tensor         ::= type_name "[" type_dim_expr { "," type_dim_expr } "]" ;
type_name           ::= NAME ;
type_dim_expr       ::= type_dim_term { ("+" | "-") type_dim_term } ;
type_dim_term       ::= type_dim_factor { ("*" | "/" | "
type_dim_factor     ::= NUMBER
                      | ".." type_name
                      | type_name
                      | "(" type_dim_expr ")"
                      ;

# -- Statements --

statement           ::= for_statement
                      | for_bind_statement
                      | if_statement
                      | scope_bind_statement
                      | return_statement
                      | yield_statement
                      | bind_statement
                      ;

for_statement       ::= "for" [ for_scope ] NAME "<-" range_expr
                        [ for_step ] [ for_carry ]
                        ( "do" suite | bind_expr )
                      ;
for_bind_statement  ::= target_list "<-" "for" [ for_scope ] NAME "<-" range_expr
                        [ for_step ] [ for_carry ]
                        ( "do" suite | bind_expr )
                      ;
for_scope           ::= "@" scoped_name ;
for_step            ::= "step" "=" expr ;
for_carry           ::= "carry" "(" target_list ")" ;

if_statement        ::= "if" expr "then" "do" suite
                        "else" "do" suite
                      ;

scope_bind_statement ::= target_list "<-" "scope" scope_ref
                          { scope_bind_kwarg } "do" suite
                        ;
scope_bind_kwarg    ::= NAME "=" kwarg_value ;
scope_ref           ::= path_lit
                      | [ "@" ] scoped_name
                      | [ "@" ] STRING
                      ;
scoped_name         ::= scoped_part { "." scoped_part } ;
scoped_part         ::= NAME | TEMPLATE_NAME ;

return_statement    ::= "return" expr_list ;
yield_statement     ::= "yield" expr_list ;
bind_statement      ::= target_list "<-" expr ;

target_list         ::= NAME { "," NAME } ;
expr_list           ::= expr { "," expr } ;

# -- Suites (statement blocks) --

suite               ::= inline_suite [ block_suite ]
                      | block_suite
                      ;
inline_suite        ::= inline_statement { ";" inline_statement } [ ";" ] ;
inline_statement    ::= return_statement | yield_statement | bind_statement ;
block_suite         ::= INDENT { statement_line NEWLINE } DEDENT ;
statement_line      ::= statement { ";" statement } [ ";" ] ;

# -- Range expressions --
#   [a..b)  half-open:     from=a, to=b       (default)
#   (a..b]  open-closed:   from=a+1, to=b+1
#   [a..b]  closed-closed: from=a,   to=b+1

range_expr          ::= range_start expr ".." expr range_end ;
range_start         ::= "[" | "(" ;
range_end           ::= "]" | ")" ;

# -- Expression hierarchy --
#   Precedence (low to high):
#     ascribe >> bind (>>=) >> pipe (|>) >> ternary/if >> or >> and >> cmp
#     >> add >> mul >> application (bare-call) >> atom
#
# Multi-line variants of if/ternary (with INDENT/DEDENT around branch
# values) and multi-line pipe/bind chains (with nl_gap newlines) are
# also accepted; see grammar.py for details.

expr                ::= expr_core [ "::" type_expr ] ;
expr_core           ::= do_expr | bind_expr ;
do_expr             ::= "do" suite ;

bind_expr           ::= pipe_expr [ ">>=" lambda_expr ] ;
lambda_expr         ::= "\" NAME "->" expr ;

pipe_expr           ::= ternary_expr { "|>" ternary_expr } ;

ternary_expr        ::= if_expr
                      | or_expr "?" tuple_value ":" tuple_value
                      | or_expr
                      ;
if_expr             ::= "if" expr "then" tuple_value "else" tuple_value
                      | or_expr
                      ;

or_expr             ::= and_expr { "or" and_expr } ;
and_expr            ::= cmp_expr { "and" cmp_expr } ;
cmp_expr            ::= add_expr { CMP_OP add_expr } ;
add_expr            ::= mul_expr { ("+" | "-") mul_expr } ;
mul_expr            ::= app_expr { ("*" | "/" | "

app_expr            ::= NAME bare_arg { bare_arg }
                      | atom
                      ;
bare_arg            ::= NAME "=" arg_expr       # keyword argument
                      | arg_expr                # positional argument
                      ;
# arg_expr mirrors the full expr precedence hierarchy (or/and/cmp/add/mul/
# if/ternary/do/ascribe) but excludes bare tuple expressions at the atom
# level -- use "(expr)" for parenthesized single-value arguments.
# See grammar.py (arg_expr through arg_mul) for the parallel hierarchy.
kwarg_value         ::= arg_expr ;

tuple_value         ::= expr { "," expr } [ "," ] ;

atom                ::= tuple_expr
                      | "(" expr ")"
                      | list_expr
                      | literal
                      | NAME
                      ;
tuple_expr          ::= "(" expr "," expr { "," expr } [ "," ] ")" ;
list_expr           ::= "[" [ expr { "," expr } [ "," ] ] "]" ;

# -- Literals --

literal             ::= NUMBER | "true" | "false" | "null" | STRING | path_lit ;
path_lit            ::= PATH_LIT ;

# -- Terminals --

NAME                ::= /[A-Za-z_](?:[A-Za-z0-9_:@]|\.(?!\.))*/
                      # Greedy: matches ns::op, mod@path, Act.swiglu as one token.
                      ;
TEMPLATE_NAME       ::= /\{[A-Za-z_][A-Za-z0-9_]*\}/ ;
NUMBER              ::= /-?[0-9]+(\.[0-9]+)?([eE][+-]?[0-9]+)?/ ;
STRING              ::= /"([^"\\]|\\.)*"|'([^'\\]|\\.)*'/ ;
PATH_LIT            ::= /@@?'([^'\\]|\\.)*'|@@?([A-Za-z_][A-Za-z0-9_]*|[0-9]+)(\.([A-Za-z_][A-Za-z0-9_]*|[0-9]+))*/
                      # @@abs.path  @rel.path  @'escaped path'  @@'abs escaped'
                      ;
CMP_OP              ::= "==" | "!=" | "<=" | ">=" | "<" | ">" ;
NEWLINE             ::= /(\r?\n[ \t]*)+/ ;
COMMENT             ::= /--[^\n]*/ ;
INDENT              ::= /* emitted by indentation post-lexer */ ;
DEDENT              ::= /* emitted by indentation post-lexer */ ;

\end{lstlisting}
\twocolumn

\end{document}